%% file: main.tex
\documentclass[letterpaper]{article} 
\usepackage[preprint]{aaai2027}  
\usepackage[hyphens]{url}  
\usepackage{graphicx} 
\usepackage{natbib}  
\usepackage{caption} 
\usepackage{algorithm}
\usepackage{algorithmic}

\usepackage{newfloat}
\usepackage{listings}
\DeclareCaptionStyle{ruled}{labelfont=normalfont,labelsep=colon,strut=off} 
\floatstyle{ruled}
\newfloat{listing}{tb}{lst}{}
\floatname{listing}{Listing}

\usepackage{booktabs}

\usepackage{amssymb}
\usepackage{amsmath}
\usepackage{xcolor}
\definecolor{blue}{RGB}{117, 159, 204} 
\definecolor{green}{RGB}{164, 185, 91} 
\definecolor{pink}{RGB}{220, 108, 110} 
\definecolor{yellow}{RGB}{207, 146, 47}
\definecolor{purple}{RGB}{124, 98, 162}
\usepackage{multirow}

\usepackage{pifont}
\newcommand{\cmark}{\ding{51}}
\newcommand{\xmark}{\ding{55}}
\usepackage[most]{tcolorbox}

\title{DepthArb: Training-Free Depth-Arbitrated Generation for \\Occlusion-Robust Image Synthesis}
\author{
Hongjin Niu\textsuperscript{\rm 1}\equalcontrib,
Jiahao Wang\textsuperscript{\rm 1}\equalcontrib,
Xirui Hu\textsuperscript{\rm 1},
Weizhan Zhang\textsuperscript{\rm 1}\corresponding,\\
Lan Ma\textsuperscript{\rm 2},
Yuan Gao\textsuperscript{\rm 2},
Feng Lei\textsuperscript{\rm 2}
}

\affiliations{
\textsuperscript{\rm 1}School of Computer Science and Technology, Xi'an Jiaotong University\\
\textsuperscript{\rm 2}China Telecom\\
\{niuhongjin,jiahaowang\}@stu.xjtu.edu.cn\\
zhangwzh@xjtu.edu.cn
}

\begin{document}

\maketitle


\begin{abstract}

Text-to-image models often struggle to synthesize correct occlusion relationships among multiple objects, especially in densely overlapping regions. Many training-free layout-guided methods enforce 2D spatial constraints but do not explicitly resolve depth-dependent attention competition, which can cause concept mixing and implausible occlusion. To address this problem, we propose DepthArb, a training-free framework that formulates occlusion generation as attention arbitration within a unified denoising trajectory. DepthArb employs two core occlusion-control mechanisms: Attention Arbitration Modulation suppresses background-object attention within foreground support according to relative depth, while Spatial Compactness Control limits attention dispersion to preserve object coherence. Because interference varies during generation, Occlusion Conflict Estimation constructs a shared spatial conflict field to adaptively weight both objectives. Through a unified spatial-text attention interface, DepthArb operates on U-Net cross-attention and the image-to-text component of MMDiT joint attention without model retraining. We further introduce OcclBench, a benchmark with continuous relative-depth specifications and occlusion-specific evaluation metrics. Experiments on OcclBench and public benchmarks show that DepthArb improves several layout and occlusion metrics over the evaluated baselines while maintaining competitive text-image alignment.

\end{abstract}


\section{Introduction}
\label{sec:intro}

\begin{figure}[t]
    \centering
    \includegraphics[width=\columnwidth]{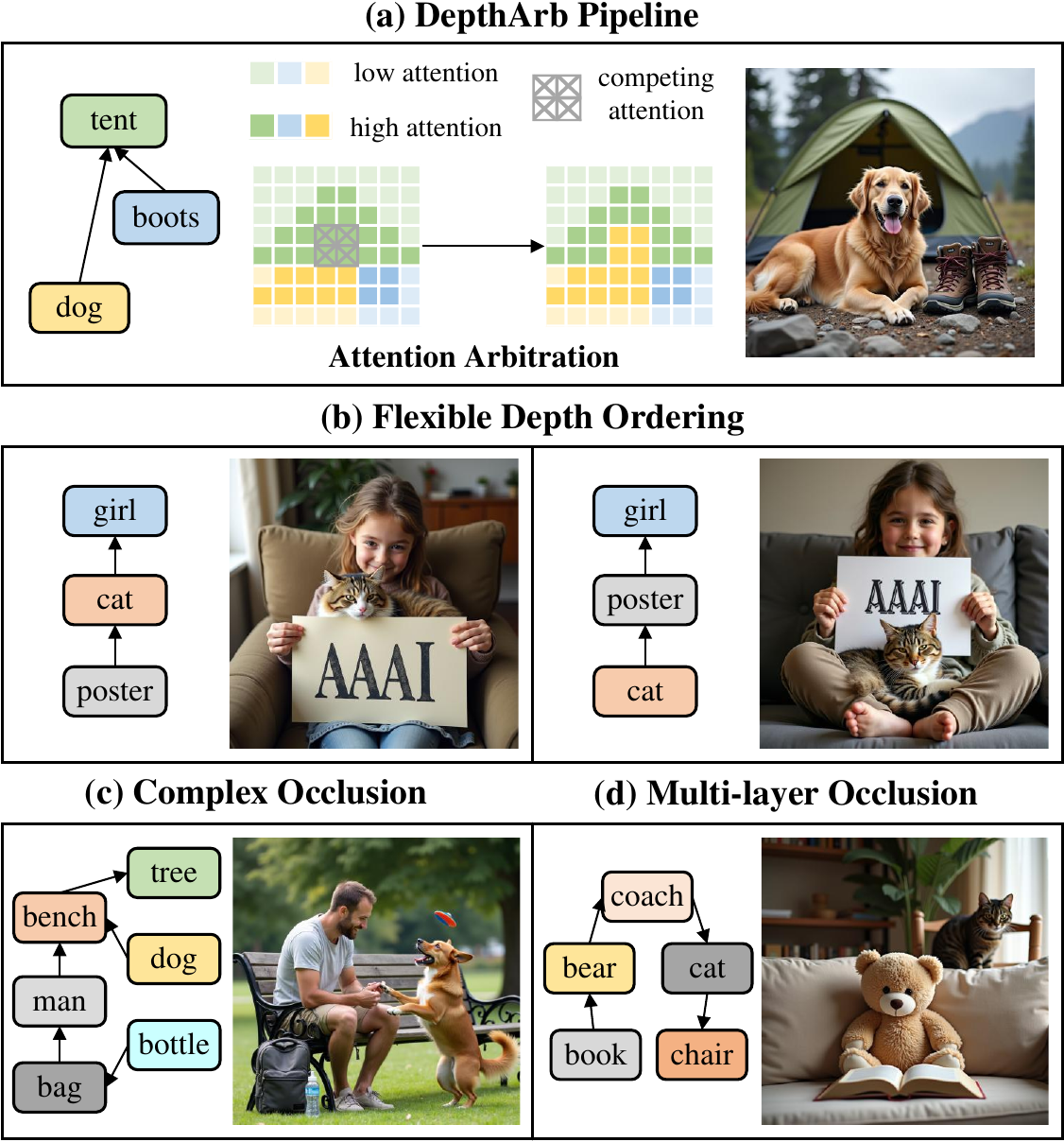}
    \caption{Our training-free framework resolves attention competition through attention arbitration, enabling flexible depth ordering, robust generation in complex occlusion scenarios, and multi-layer occlusion synthesis.}
    \label{fig:teaser}
\end{figure}

\nocite{oneactor,echoshot,echomotion,dynamicid,motionweaver,schedule}

Diffusion- and flow-based models \cite{ho2020denoising, song2021denoising, sohl2015deep, esser2024scaling, labs2025flux1kontextflowmatching} have become dominant paradigms in image synthesis across U-Net-based architectures and recent MMDiT backbones. However, complex multi-object composition remains challenging, as modern text-to-image models \cite{ho2021classifierfree, rombach2022high, podell2023sdxlimprovinglatentdiffusion} frequently exhibit spatial inconsistency. Existing spatial controls primarily employ attention-based guidance \cite{xie2023boxdiff,chen2024training, ICLR2024_6f61f25a, chefer2023attend, kim2022dag, kim2023dense,feng2023trainingfree,hertz2022prompttopromptimageeditingcross} or region-aware generation \cite{BarTal2023MultiDiffusionFD, ohanyan2024zero,Zhan_2025_ICCV, jimenez2023mixture, ma2024directed, zheng2023layoutdiffusion, wang2024compositional, lee2023syncdiffusion, gafni2022make}, substantially improving alignment between textual prompts and prescribed 2D layouts.


Despite these advances, many layout-guided pipelines specify where objects should appear but do not explicitly represent depth hierarchy or occlusion ordering. Providing a visibility order resolves the desired foreground-background relation, but it does not by itself prevent the corresponding object representations from competing for the same spatial support during joint denoising. Such competition can cause background attention to intrude into foreground regions, leading to attention hijacking, concept mixing, or missing objects \cite{balaji2022ediff, chefer2023attend}. Existing order-aware methods \cite{liang2025vodiff, Zhan_2025_ICCV, Chen_2026_CVPR} handle this ambiguity through sequential object-wise denoising, latent-layer rendering, or region-specific generation branches, but at the cost of separate rendering stages.

To resolve attention competition without decomposing the generation path, we introduce \textbf{DepthArb}, a training-free framework that performs conflict-adaptive attention arbitration within a unified denoising trajectory. Figure~\ref{fig:teaser} illustrates how DepthArb arbitrates competing attention to support flexible depth ordering and complex multi-layer occlusion. DepthArb integrates two core mechanisms: Attention Arbitration Modulation (\textbf{AAM}) and Spatial Compactness Control (\textbf{SCC}). AAM imposes depth-aware orthogonality to suppress background attention within foreground support, rather than rendering objects through separate ordered stages. SCC complements this arbitration by minimizing spatial variance around the attention-weighted centroid, preventing object attention from becoming diffuse after redistribution. An occlusion conflict signal adaptively focuses these constraints on ambiguous interactions. Operating on a unified spatial-text attention representation, DepthArb supports both U-Net cross-attention and MMDiT joint attention. To evaluate the effectiveness of our method, we establish \textbf{OcclBench}, a benchmark tailored for occlusion-aware generation with contextually grounded object groups and specialized metrics for depth ordering and semantic integrity. Extensive experiments on both OcclBench and standard benchmarks demonstrate the effectiveness of our method. The accurate occlusion handling, sharp boundaries, and high-fidelity text-image alignment confirm that DepthArb successfully resolves complex spatial compositions.

To summarize, the key contributions of our work are as follows:
\begin{itemize}
    \item We identify intrinsic attention competition as a critical challenge in multi-object depth-ordered composition and characterize how overlapping layouts trigger background leakage into foreground regions.
    
    \item We propose DepthArb, a training-free framework that preserves simultaneous generation and resolves spatial conflicts via Attention Arbitration Modulation and Spatial Compactness Control. Its unified attention interface supports both U-Net and MMDiT backbones without sequential object-wise rendering.

    \item We introduce OcclBench, a new benchmark with specialized metrics to evaluate foreground-background relations. Evaluations on OcclBench and public benchmarks show improvements over the evaluated baselines on several depth-ordering and visual-fidelity metrics.
\end{itemize}

\section{Related Work}
\subsection{Training-Free Spatial Guidance}
To align image synthesis with user-defined spatial constraints without the computational burden of fine-tuning, training-free spatial guidance \cite{xie2023boxdiff, chen2024training, ICLR2024_6f61f25a, ohanyan2024zero, Zhan_2025_ICCV, lee2024groundit, regionprompt,chefer2023attend,kim2023dense, wang2025spotactor} has emerged as a dominant research paradigm. Current approaches typically fall into two categories: gradient-based optimization and region-based fusion. Gradient-based methods \cite{xie2023boxdiff,chen2024training,ICLR2024_6f61f25a} define energy functions that penalize cross-attention activations falling outside target bounding boxes \cite{hertz2022prompttopromptimageeditingcross}. While effective for spatially disjoint objects, overlapping boxes can introduce competing guidance signals because multiple object objectives act on the same spatial support, which may lead to texture contamination and semantic entanglement \cite{feng2023trainingfree}. Conversely, region-based methods \cite{ohanyan2024zero,Zhan_2025_ICCV} denoise regional predictions independently and blend them into a global canvas. Although effective at separating disjoint regions, blending independently generated features in overlapping areas can create ambiguous object ownership without an explicit visibility rule \cite{ohanyan2024zero}. Moreover, most attention-guided spatial objectives are formulated around U-Net cross-attention, whereas MMDiT backbones encode image-text interactions through joint attention. DepthArb addresses these limitations through a shared spatial-text interface and conflict-adaptive arbitration within a unified latent trajectory.

\subsection{Occlusion-Aware Synthesis}
Achieving realistic multi-object composition \cite{li2023gligen, BarTal2023MultiDiffusionFD} requires extending control from 2D positioning to relative depth ordering. Adapter-based frameworks \cite{mou2024t2i, zhang2023adding} provide geometric control through dense, pixel-aligned depth conditions, but require additional conditioning modules and offer less flexibility than sparse box-and-depth specifications. Recent approaches \cite{Zhan_2025_ICCV,Chen_2026_CVPR,li2026occlusionformer, agrawal2026seethrough3d} address overlap through volume rendering, layer-wise binding, or explicit z-order arrangement. Closest to our setting, VODiff \cite{liang2025vodiff} assigns objects to discrete visibility layers, renders them through sequential object-wise denoising stages, and optimizes attention over pre-partitioned overlapping, visible, and background regions. In contrast, DepthArb retains simultaneous generation in a unified latent trajectory and uses continuous depth to modulate conflict-adaptive orthogonality and spatial compactness. 

\section{Method}
\subsection{Overview}
\begin{figure*}[tb]
  \centering
  \includegraphics[width=\textwidth]{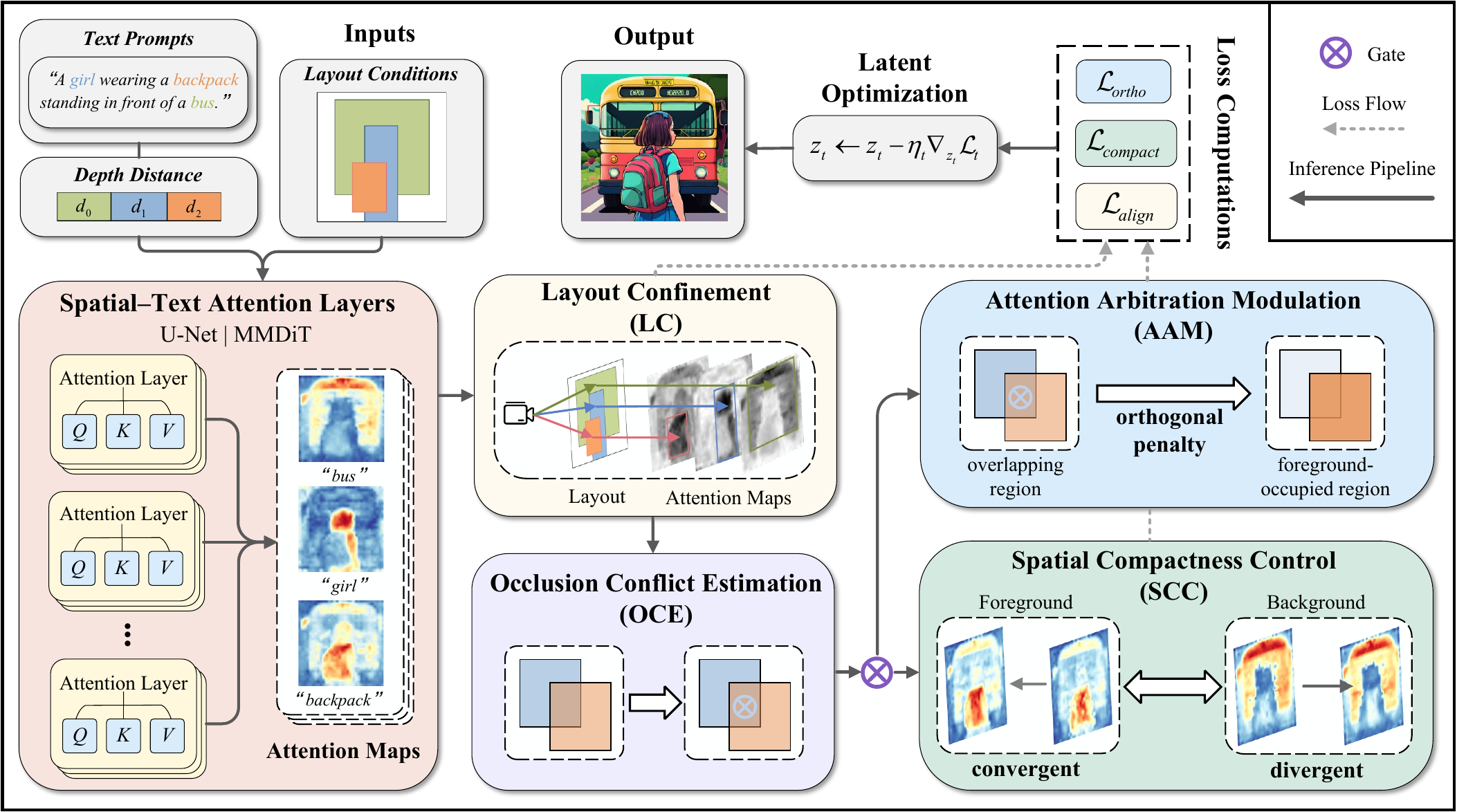}
  \caption{The overall architecture of DepthArb.
The framework converts U-Net cross-attention or the image-to-text component of MMDiT joint attention into a unified spatial-text representation. Layout Confinement establishes spatial alignment, while Occlusion Conflict Estimation adaptively reweights the depth-aware orthogonality of Attention Arbitration Modulation and the compactness regularization of Spatial Compactness Control.
  }
  \label{fig:method}
\end{figure*}
In our task, users provide a textual prompt and, for each object $i$, a textual phrase with its token set $\mathcal{T}_i$, a bounding box $\mathbf{B}_i$, and a relative depth $d_i$ representing its distance to the camera, together with a user given ordered occlusion set $\mathcal{O}$ of foreground-background pairs $(i,j)$. Because the input depths may follow arbitrary scales, we normalize them within each scene as $\bar{d}_i=(d_i-d_{\min})/(d_{\max}-d_{\min}+\epsilon)$, where $d_{\min}=\min_k d_k$, $d_{\max}=\max_k d_k$, and $\epsilon$ ensures numerical stability. Thus, $\bar{d}_i\in[0,1]$, with smaller values denoting nearer objects, and $\bar{d}_i<\bar{d}_j$ for each foreground--background pair $(i,j)\in\mathcal{O}$. Our goal is to satisfy the specified layouts and depth ordering while preserving the visual coherence of partially occluded objects.

DepthArb preserves a unified denoising trajectory in which all objects are generated jointly. As illustrated in Figure~\ref{fig:method}, its guidance objectives operate on a shared spatial--text attention representation, enabling the same training-free framework to support both U-Net cross-attention and MMDiT joint attention.

\subsection{Layout Confinement via Attention Guidance}
\label{sec:lc}
We first convert backbone-specific attention into a common spatial-text representation. For U-Net backbones with cross-attention, we extract attention maps from the conditional branch and aggregate the element-wise top 25\% responses across heads. For MMDiT backbones with joint attention, we extract the image-query-to-text-key submatrix and average it across heads. The resulting attention maps are reshaped to their corresponding spatial resolutions and used by the same guidance objectives. We then establish spatial fidelity by confining each object's semantic representation within its designated bounding box.

Formally, for each object $i$, we average the attention maps associated with its textual tokens to obtain $\mathbf{A}_i \in \mathbb{R}^{H \times W}$. Given its bounding box $\mathbf{B}_i$, we define a binary mask $\mathbf{M}_i \in \{0,1\}^{H \times W}$ that equals 1 inside the box and 0 elsewhere. The fraction of object attention localized within the prescribed region is
\begin{equation}
f_i = \frac{\sum_{x,y}\mathbf{A}_i(x,y)\mathbf{M}_i(x,y)}{\sum_{x,y}\mathbf{A}_i(x,y)+\epsilon}.
\end{equation}
The ratio $f_i$ reflects how well the semantic activation for object $i$ is localized within its prescribed region.

The layout confinement objective generally minimizes the overall deviation of the alignment ratio. To account for perspective geometry, we modulate this constraint using the normalized object depth $\bar{d}_i$. Since closer objects have higher visibility priority and perceptual dominance in occlusion relationships, attention leakage beyond their designated boundaries is perceptually more disruptive. Accordingly, we use $1-\bar{d}_i$ as a proximity weight and incorporate it into the alignment loss:
\begin{equation}
\mathcal{L}_{\mathrm{align}} = \sum_i (1-\bar{d}_i) \cdot \bigl(1 - f_i\bigr)^2.
\end{equation}

During the generation process, this objective functions as a training-free guidance signal that is applied at each active guidance step. Its gradient steers the generation to suppress attention leakage beyond designated boundaries, encouraging each object to remain spatially anchored at the latent feature level.

Whereas depth specifies the desired ordering, we next introduce \textbf{Occlusion Conflict Estimation} (\textbf{OCE}), which constructs a shared spatial conflict field from the current attention state. For each selected layer $\ell\in\mathcal{L}$, we normalize each object map as $\hat{\mathbf{A}}_i^{(\ell)}=\mathbf{A}_i^{(\ell)}/(\max_{x,y}\mathbf{A}_i^{(\ell)}(x,y)+\epsilon)$. Let $\mathbf{R}^{(\ell)}$ be the priority map obtained by rasterizing $r_i=1-\bar d_i$ within each box, with higher-priority objects overwriting lower-priority ones in overlapping regions. We define the pairwise soft intersection $m_{ij}^{(\ell)}=\min(\hat{\mathbf{A}}_i^{(\ell)},\hat{\mathbf{A}}_j^{(\ell)})$, the attention-weighted priority $\rho_i^{(\ell)}=\langle\mathbf{A}_i^{(\ell)},\mathbf{R}^{(\ell)}\rangle/(\|\mathbf{A}_i^{(\ell)}\|_1+\epsilon)$, and the normalized priority-boundary strength $G^{(\ell)}=\operatorname{Norm}(\|\nabla_S\mathbf{R}^{(\ell)}\|_2)$. OCE combines attention overlap, priority ambiguity, and geometric complexity as
\begin{equation}
\begin{aligned}
O^{(\ell)} &= \max_{i<j}m_{ij}^{(\ell)}, \\
D^{(\ell)} &= \max_{i<j}e^{-|\rho_i^{(\ell)}-\rho_j^{(\ell)}|}m_{ij}^{(\ell)}, \\
\mathbf{C}^{(\ell)} &= \sigma\!\left(O^{(\ell)}+D^{(\ell)}+G^{(\ell)}\right).
\end{aligned}
\end{equation}
Here $\nabla_S$ is the Sobel operator and $\operatorname{Norm}$ denotes spatial min--max normalization. Layout Confinement (LC) remains independent of OCE; the conflict field is used only to modulate the subsequent arbitration and compactness objectives. Thus, OCE provides state-dependent modulation without requiring a fixed region partition.

\subsection{Attention Arbitration Modulation}
\label{sec:aam}
In scenarios where multiple objects occupy partially overlapping spatial regions, independently enforcing layout constraints is often insufficient to prevent attention interference. Specifically, background attention may propagate into foreground areas even when both objects individually satisfy their layout constraints, producing transparency artifacts or ambiguous depth ordering. Unlike region-wise visibility objectives that explicitly amplify foreground attention in predefined visible or overlapping regions, \textbf{Attention Arbitration Modulation} applies a unidirectional, depth-aware orthogonality constraint to background attention within foreground support. This resolves competition without sequential object rendering or direct foreground amplification.

Given an occlusion pair $(i,j)$, where object $i$ is designated as the foreground and object $j$ as the background, we first aggregate the attention responses associated with object $j$ across heads and token channels, yielding a spatial attention map $\mathbf{A}_j^{(\ell)} \in \mathbb{R}^{H \times W}$ at layer $\ell$. To quantify attention interference in overlapping regions, we define the overlap mask as $\mathbf{M}_{ij}=\mathbf{M}_i\odot\mathbf{M}_j$ and measure the degree to which background attention intrudes into this region via a mask-normalized overlap serving as an orthogonality surrogate:
\begin{equation}
\mathcal{I}_{i \leftarrow j}^{(\ell)}
=
\frac{
\sum_{x,y} \mathbf{A}_j^{(\ell)}(x,y)\, \mathbf{M}_{ij}(x,y)
}{
\sum_{x,y} \mathbf{M}_{ij}(x,y) + \epsilon
},
\end{equation}
where $\mathcal{I}_{i \leftarrow j}^{(\ell)}$ captures the extent to which background attention aligns with foreground-occupied regions, which primarily correspond to overlapping or competing spatial areas under occlusion. Minimizing this term encourages the background attention $\mathbf{A}_j^{(\ell)}$ to become orthogonal to the overlap mask $\mathbf{M}_{ij}$. This effectively suppresses background features in regions that should be dominated by the foreground, thereby resolving feature-level competition.

To account for perspective effects in image formation, we further modulate this orthogonality constraint using object depth. Let $\bar{d}_i$ and $\bar{d}_j$ denote the normalized depths of the foreground and background objects, respectively. Objects closer to the camera generally exhibit stronger visual dominance and impose stricter spatial constraints on occluded objects. We therefore define the depth-aware arbitration weight as $\lambda_{ij}=\lambda_0\exp\!\left(\alpha(\bar{d}_j-\bar{d}_i)/\tau\right)$, where $\lambda_0$ is a base repulsion weight, $\alpha$ controls the sharpness of the depth-based modulation, and $\tau$ is a temperature parameter for stabilizing gradients. Since $\bar{d}_i < \bar{d}_j$ for a foreground--background pair, this formulation enforces stronger orthogonality as their relative depth separation increases, while allowing more flexible attention sharing for depth-ambiguous objects. To emphasize arbitration in conflict-prone foreground regions, we average the OCE field within the foreground mask as $c_i^{(\ell)}=\frac{\sum_{x,y}\mathbf{C}^{(\ell)}(x,y)\mathbf{M}_i(x,y)}{\sum_{x,y}\mathbf{M}_i(x,y)+\epsilon}$. Combining the above components, the attention arbitration loss is finally expressed as:
\begin{equation}
\mathcal{L}_{\mathrm{ortho}}
=
\frac{1}{|\mathcal{L}|}\sum_{\ell\in\mathcal{L}}\sum_{(i,j)\in\mathcal{O}}
\lambda_{ij}
\cdot
c_i^{(\ell)}
\cdot
\mathcal{I}_{i \leftarrow j}^{(\ell)},
\end{equation}
which uses OCE to emphasize non-orthogonal attention interactions in conflict-prone foreground regions.

During the generation process, this loss acts as a training-free guidance signal, whose gradient encourages background attention to be redistributed outside the foreground spatial region. As a result, competing objects are spatially disentangled at the latent feature level, leading to clearer depth-consistent occlusion ordering and improved compositional coherence.

\subsection{Spatial Compactness Control}
\label{sec:scc}

While LC and AAM confine object attention and redistribute competing background activations, this arbitration alone does not guarantee compact object representations. Attention may still spread excessively within a valid region, leading to visually unstable appearances. We therefore introduce \textbf{Spatial Compactness Control} to regularize the internal distribution of object-specific attention. Foreground attention is encouraged to remain spatially concentrated, whereas background attention is allowed to remain relatively diffuse, as illustrated in Figure~\ref{fig:method}.

Concretely, for each object $i$ at each selected attention layer (with the layer index omitted below for clarity), we normalize its cross-attention map $\mathbf{A}_i$ as a spatial probability distribution and compute its attention-weighted centroid:
\begin{equation}
\begin{aligned}
\tilde{\mathbf{A}}_i(x,y)
&=\frac{\mathbf{A}_i(x,y)}{\sum_{x,y} \mathbf{A}_i(x,y) + \epsilon}, \\
\boldsymbol{\mu}_i
&=\sum_{x,y}\tilde{\mathbf{A}}_i(x,y)\,\mathbf{p}(x,y).
\end{aligned}
\end{equation}
Here, $\epsilon$ is a small constant for numerical stability and $\mathbf{p}(x,y)$ denotes the normalized spatial coordinates. The spatial diffusion of attention is then quantified by the second-order moment around this mean:
\begin{equation}
\mathrm{Var}_i
=
\sum_{x,y}
\tilde{\mathbf{A}}_i(x,y)\,
\left\|
\mathbf{p}(x,y) - \boldsymbol{\mu}_i
\right\|_2^2 .
\end{equation}

A lower value of $\mathrm{Var}_i$ indicates a more compact and spatially coherent attention activation, whereas larger values correspond to increased attention divergence. Minimizing this term encourages the spatial concentration of object-specific attention and improves the stability of object appearance.

To incorporate perspective effects, we further modulate this compactness constraint using object-level depth information. Since closer objects have higher visibility priority and perceptual dominance in occlusion relationships, attention dispersion for such objects is more perceptually salient. Accordingly, we weight the compactness regularization by $1-\bar{d}_i$, enforcing stronger concentration for nearer objects while permitting looser spatial support for farther objects. We summarize the overall conflict at layer $\ell$ using the spatial mean $g^{(\ell)}=\frac{1}{HW}\sum_{x,y}\mathbf{C}^{(\ell)}(x,y)$, which increases the relative compactness weight for layers exhibiting stronger conflict. The resulting spatial compactness objective is formulated as: 
\begin{equation}
\mathcal{L}_{\mathrm{compact}}
=
\frac{1}{|\mathcal{L}|}\sum_{\ell\in\mathcal{L}}g^{(\ell)}
\sum_i(1-\bar{d}_i)\cdot\mathrm{Var}_i^{(\ell)}.
\end{equation}

During generation, this objective serves as a training-free guidance signal that suppresses excessive attention diversion at the latent level. Combined with layout confinement and attention arbitration, it promotes depth-consistent and spatially coherent object representations under complex occlusions.

\subsection{Staged Latent Optimization Pipeline}
\label{sec:inference}

We guide generation in a training-free manner by backpropagating the attention-based objectives only to the current latent representation $\mathbf{z}_t$, while keeping the pretrained model parameters fixed. All objects remain jointly represented in the same latent state; no object-specific denoising sub-stage or separately masked noise prediction is introduced. The guidance schedule is adapted to each backbone. For U-Net-based diffusion models, the initial structural stage applies LC, AAM, and SCC; the following guided stage removes AAM while retaining LC and SCC, after which denoising proceeds without latent guidance. For MMDiT-based models, all three objectives are applied within a single early guided window, followed by standard denoising without latent guidance. This design concentrates structural intervention in the coarse generation phase and preserves the backbone's native synthesis dynamics at later steps.

At each active guidance step, OCE is recomputed from the current attention maps to update the modulation weights for SCC and, when enabled, AAM. The latent variable is then iteratively refined using the resulting loss gradient:
\begin{equation}
    \mathbf{z}_t \leftarrow \mathbf{z}_t - \eta_t \nabla_{\mathbf{z}_t} \mathcal{L}_t,
\end{equation}
where $\eta_t$ denotes the latent update step size and $\mathcal{L}_t=\mathcal{L}_{\mathrm{align}}+\lambda_{\mathrm{compact}}\mathcal{L}_{\mathrm{compact}}+s_t\lambda_{\mathrm{ortho}}\mathcal{L}_{\mathrm{ortho}}$, with $s_t=1$ when AAM is active and $s_t=0$ otherwise. Each guided update is followed by the backbone-specific denoising operation: 
U-Net-based models use classifier-free guidance with an LMS step, whereas MMDiT-based models use native guidance conditioning.

\section{Experiments}
\begin{figure*}[!t]
\centering
\setlength{\tabcolsep}{1pt}      
\renewcommand{\arraystretch}{0.5} 

\begin{tabular}{cccccccc}
    \footnotesize \parbox[c]{0.12\textwidth}{\centering \textbf{Input Layout}} &
    \footnotesize \parbox[c]{0.12\textwidth}{\centering \textbf{Layout} \\ \textbf{Guidance}} &
    \footnotesize \parbox[c]{0.12\textwidth}{\centering \textbf{R\&B}} &
    \footnotesize \parbox[c]{0.12\textwidth}{\centering \textbf{VODiff}} &
    \footnotesize \parbox[c]{0.12\textwidth}{\centering \textbf{LaRender}} &
    \footnotesize \parbox[c]{0.12\textwidth}{\centering \textbf{CreatiLayout}} &
    \footnotesize \parbox[c]{0.12\textwidth}{\centering \textbf{Instance-}\\ \textbf{Assemble}} &
    \footnotesize \parbox[c]{0.12\textwidth}{\centering \textbf{Ours}} \\
    
    \addlinespace[3pt]

    \includegraphics[width=0.12\textwidth]{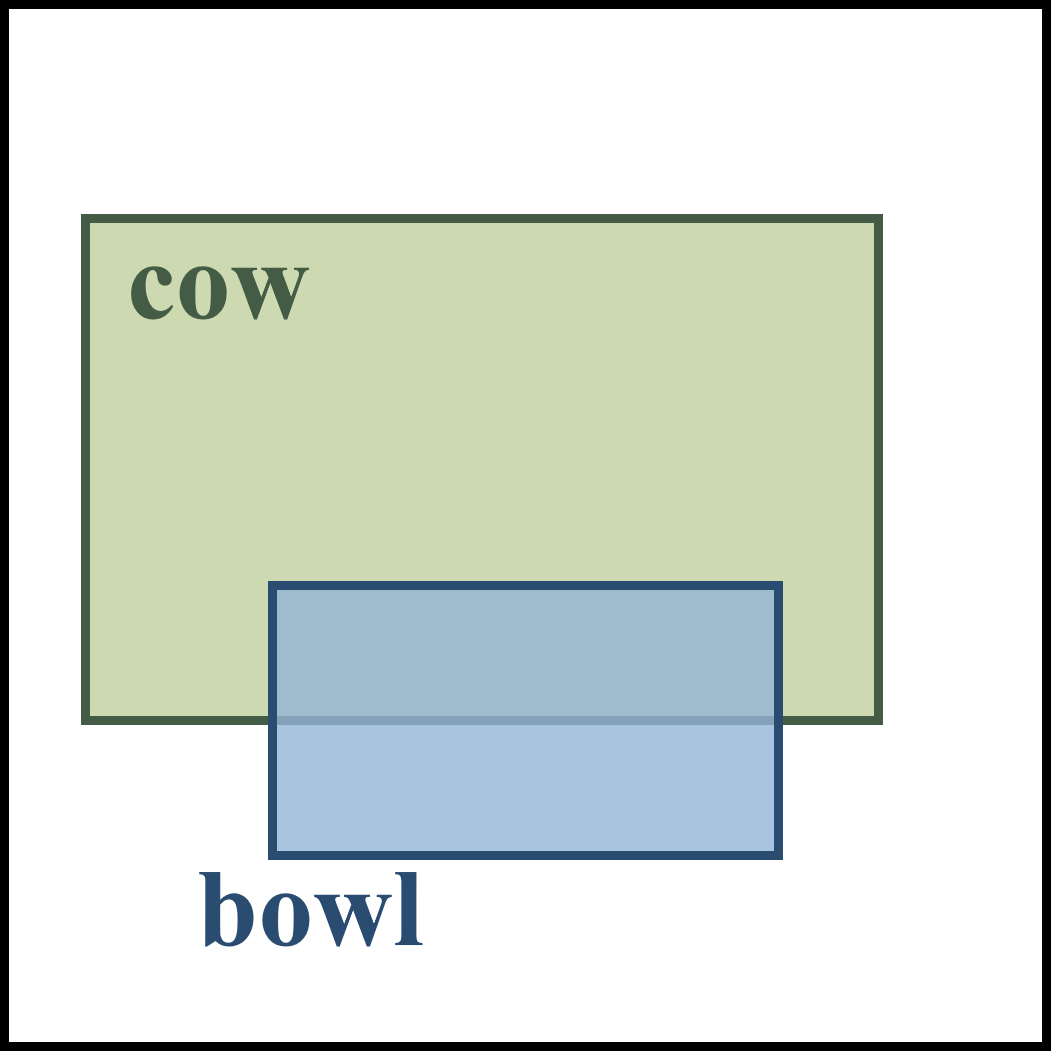} &
    \includegraphics[width=0.12\textwidth]{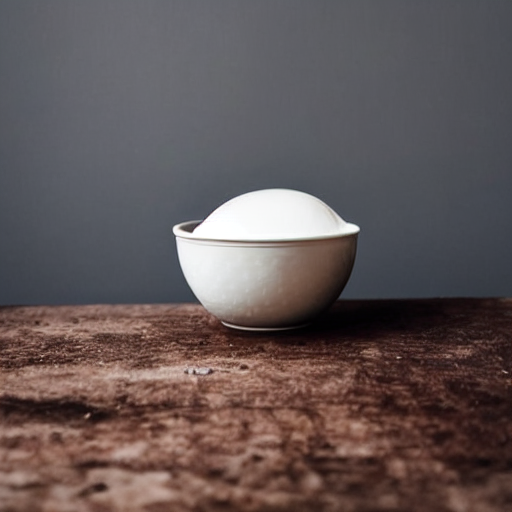} &
    \includegraphics[width=0.12\textwidth]{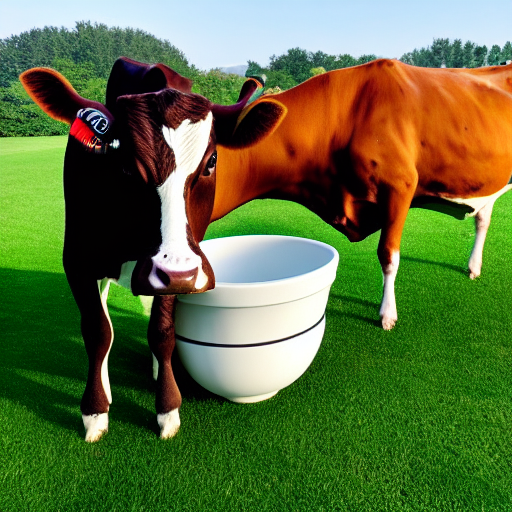} &
    \includegraphics[width=0.12\textwidth]{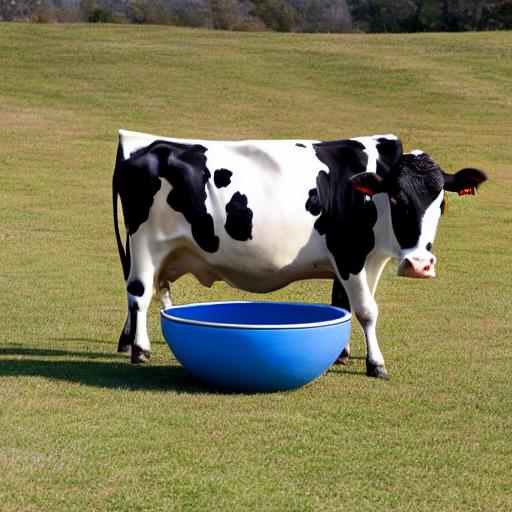} &
    \includegraphics[width=0.12\textwidth]{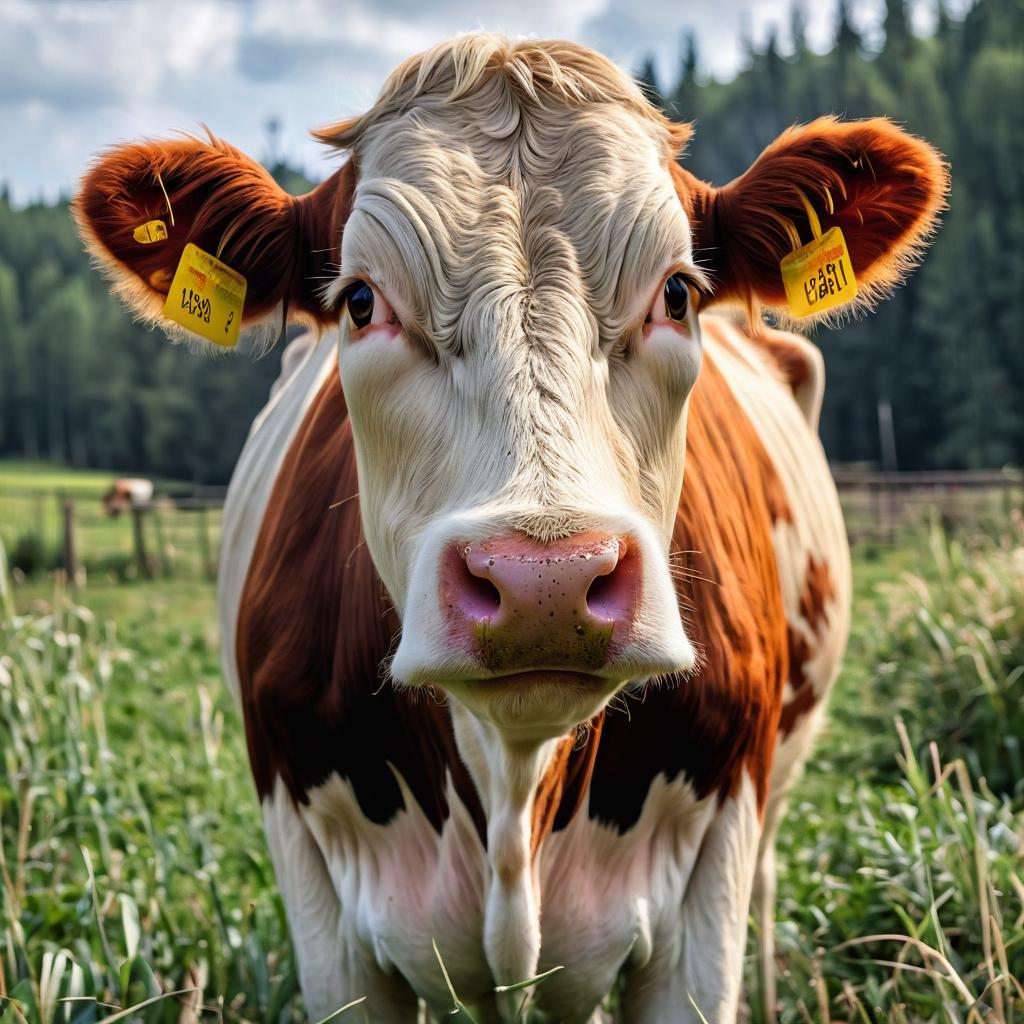} &
    \includegraphics[width=0.12\textwidth]{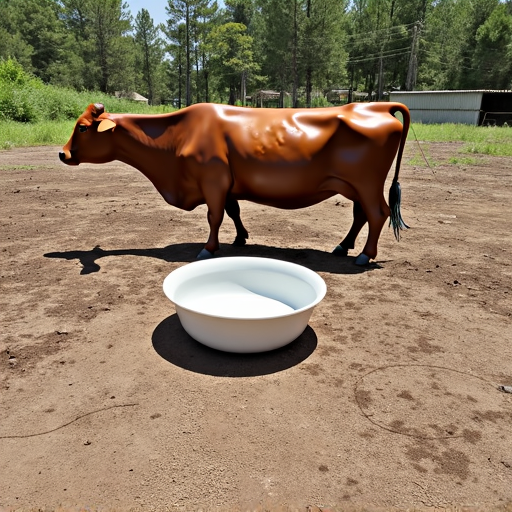} &
    \includegraphics[width=0.12\textwidth]{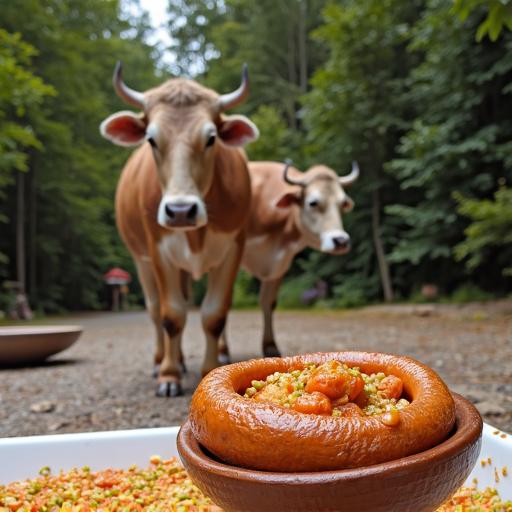} &
    \includegraphics[width=0.12\textwidth]{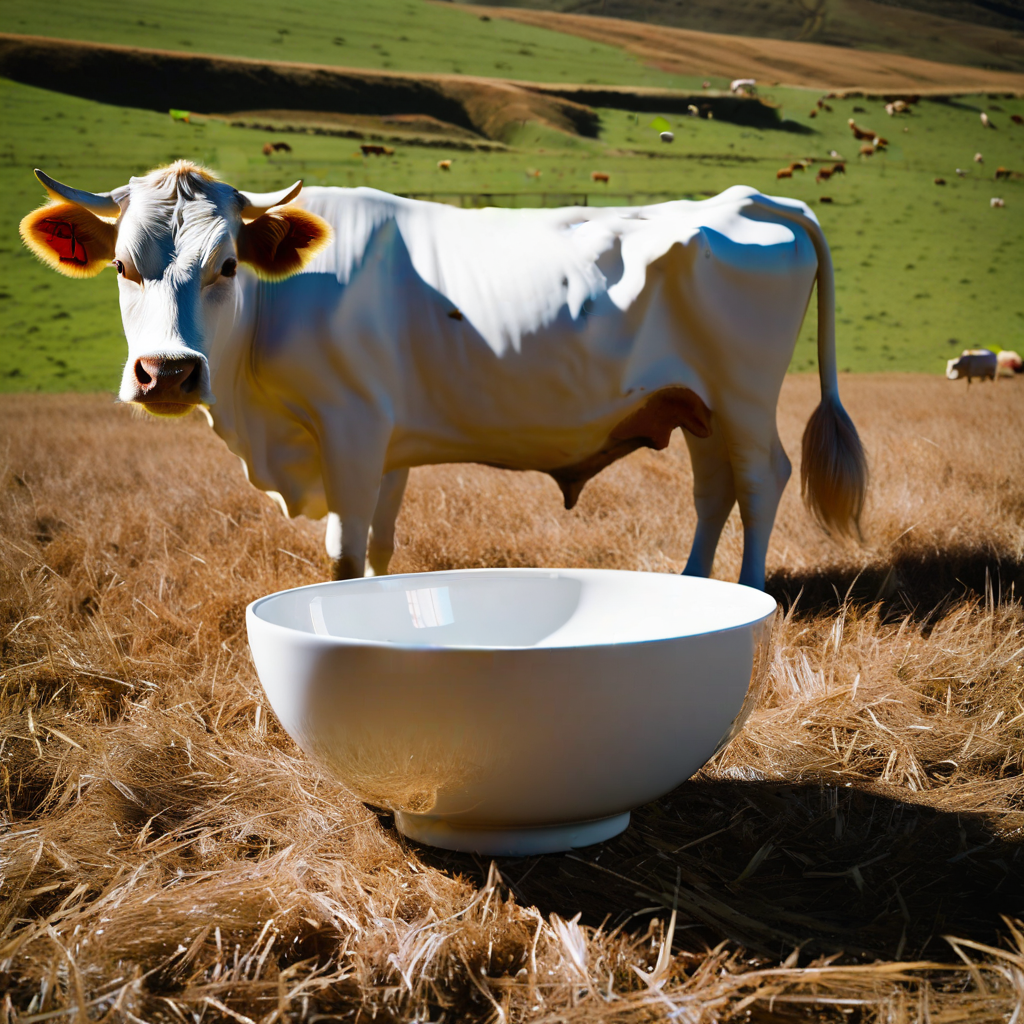} \\
    \multicolumn{8}{l}{\scriptsize \textit{Prompt: A \textcolor{green}{bowl} in front of a \textcolor{blue}{cow}.}} \\
    \addlinespace[3pt]
    
    \includegraphics[width=0.12\textwidth]{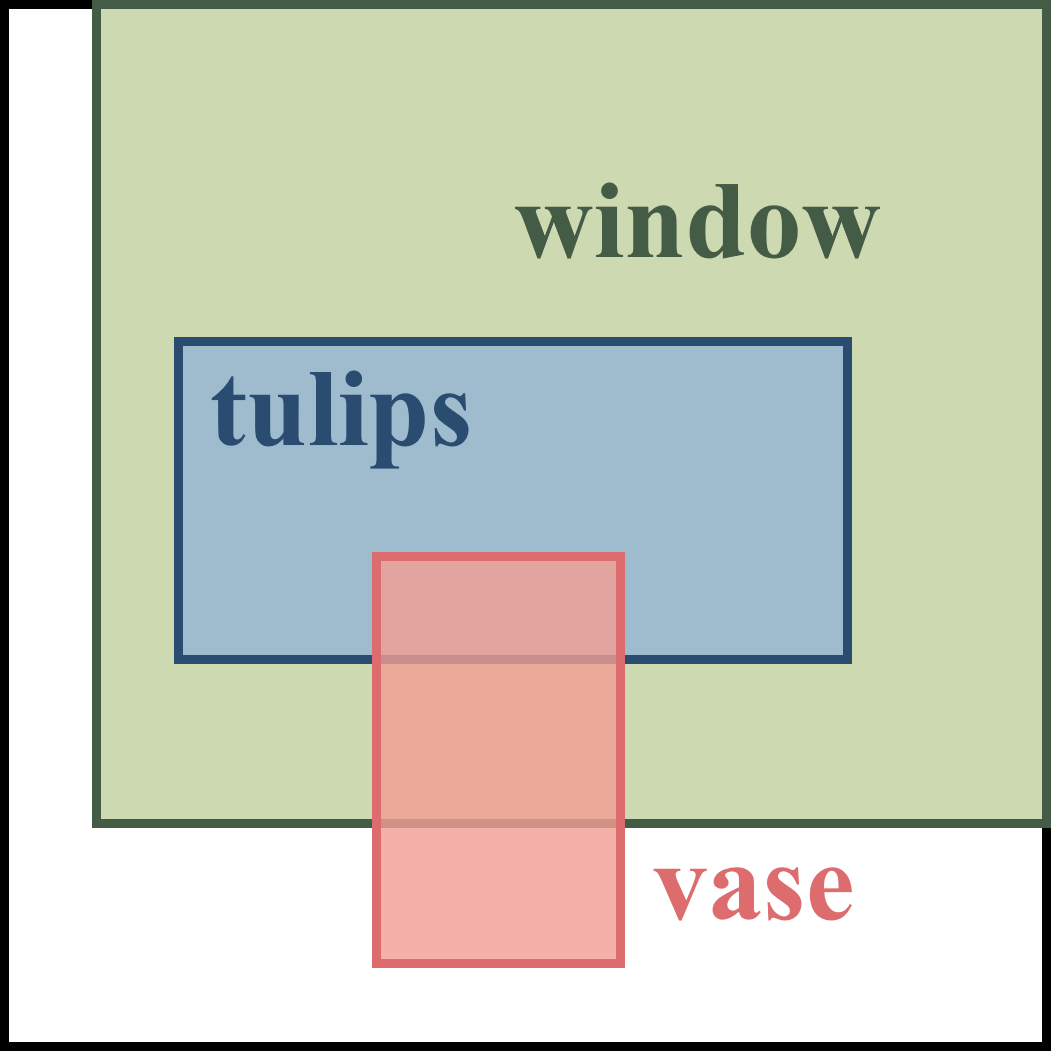} &
    \includegraphics[width=0.12\textwidth]{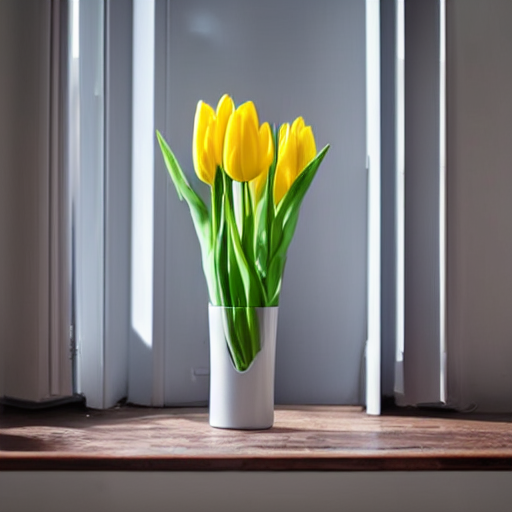} &
    \includegraphics[width=0.12\textwidth]{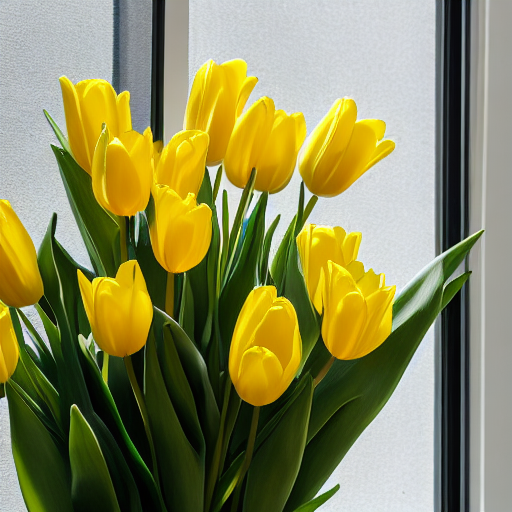} &
    \includegraphics[width=0.12\textwidth]{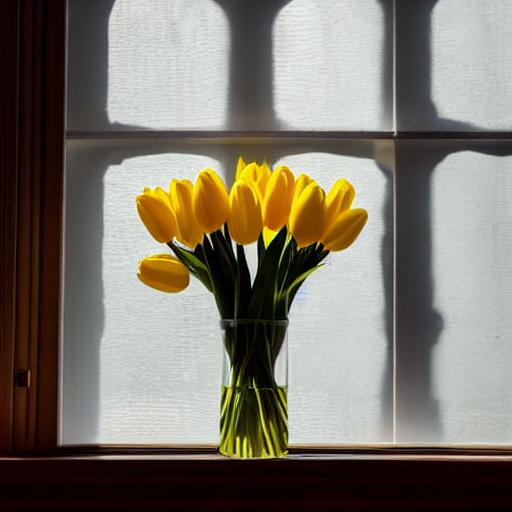} &
    \includegraphics[width=0.12\textwidth]{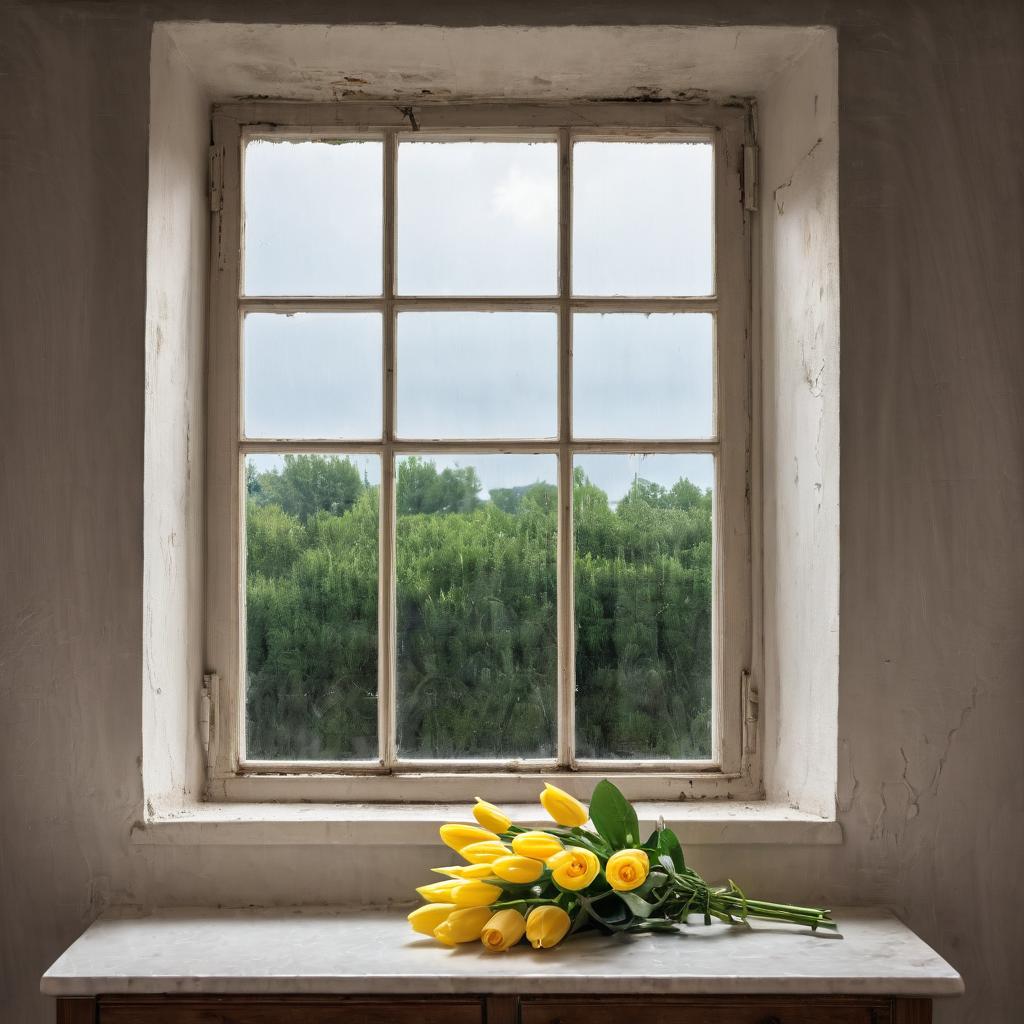} &
    \includegraphics[width=0.12\textwidth]{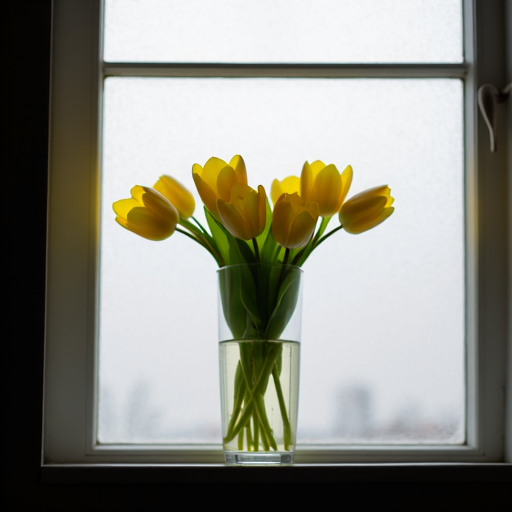} &
    \includegraphics[width=0.12\textwidth]{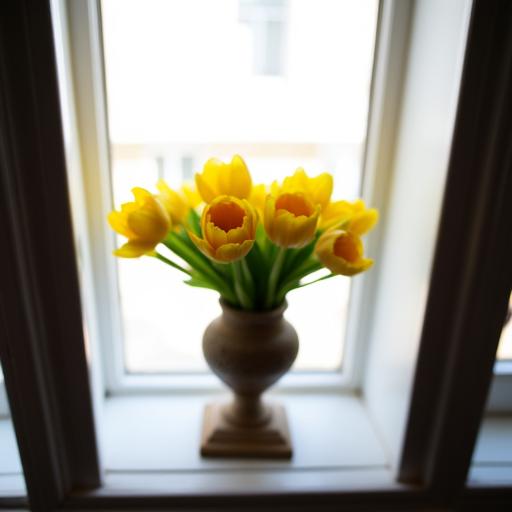} &
    \includegraphics[width=0.12\textwidth]{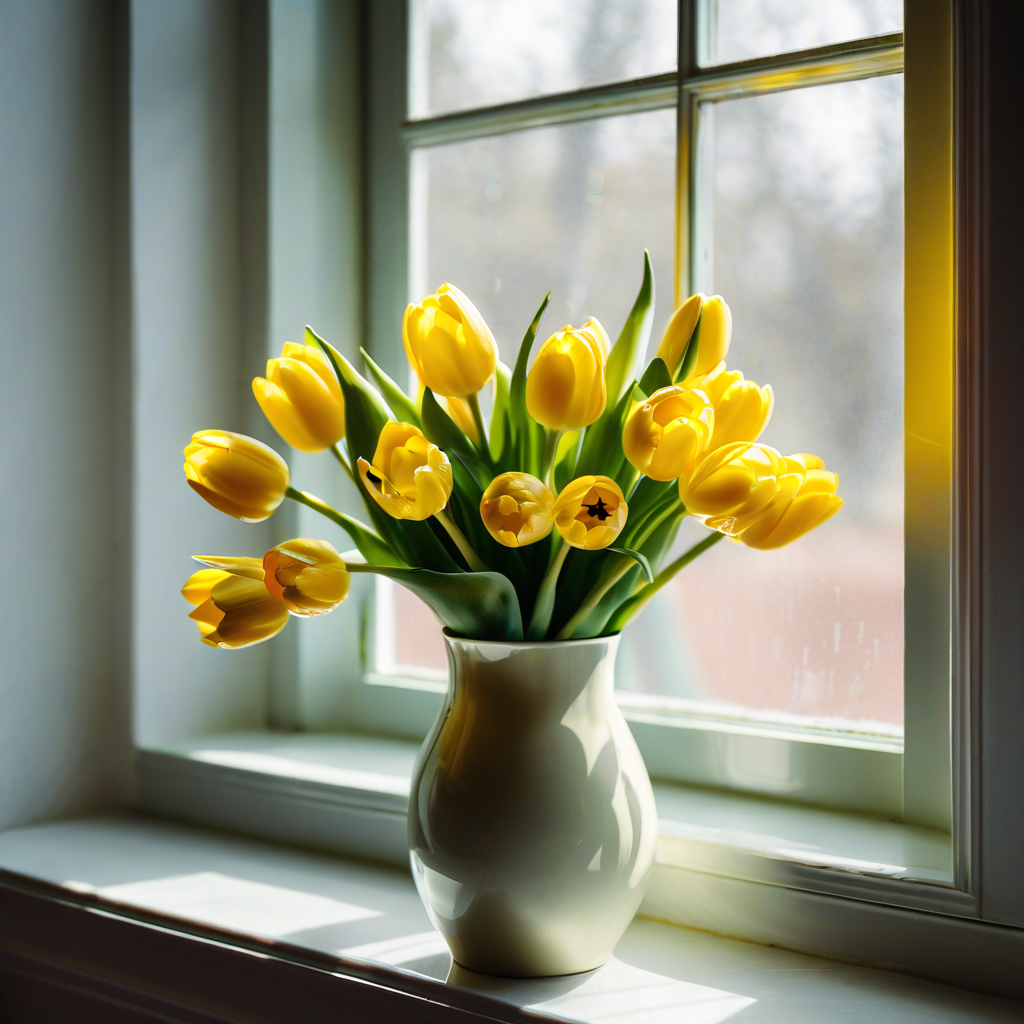} \\
    \multicolumn{8}{l}{\scriptsize \textit{Prompt: A flower \textcolor{pink}{vase} containing yellow \textcolor{blue}{tulips} standing in front of a \textcolor{green}{window}.}} \\
    \addlinespace[3pt]

    \includegraphics[width=0.12\textwidth]{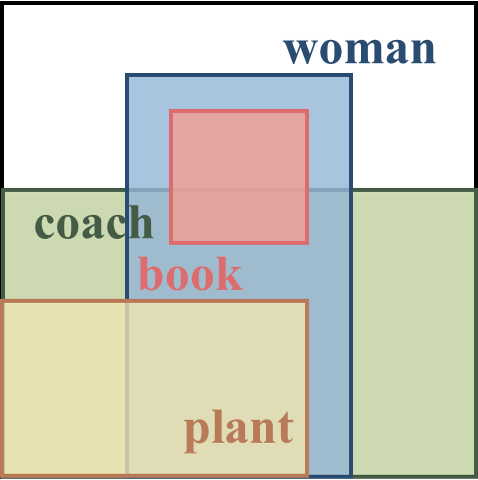} &
    \includegraphics[width=0.12\textwidth]{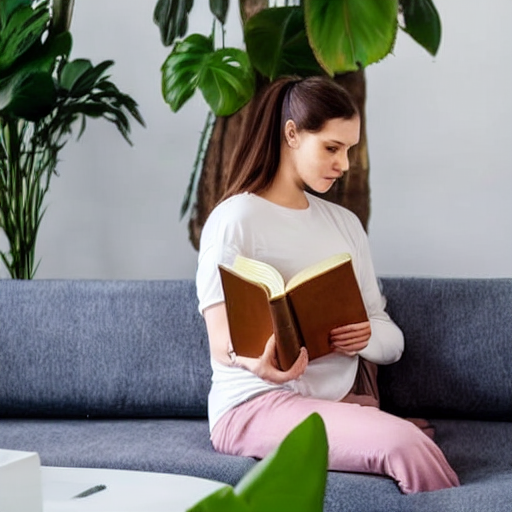} &
    \includegraphics[width=0.12\textwidth]{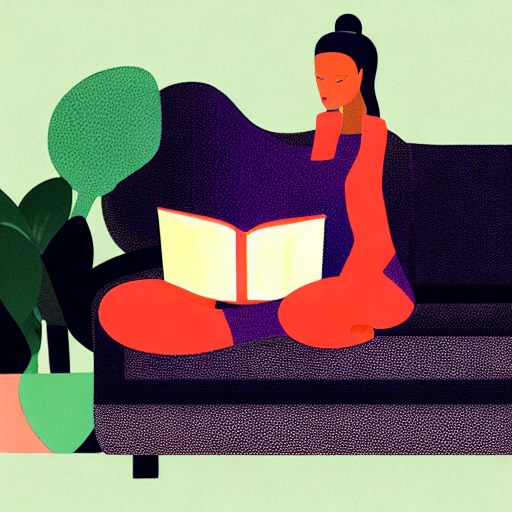} &
    \includegraphics[width=0.12\textwidth]{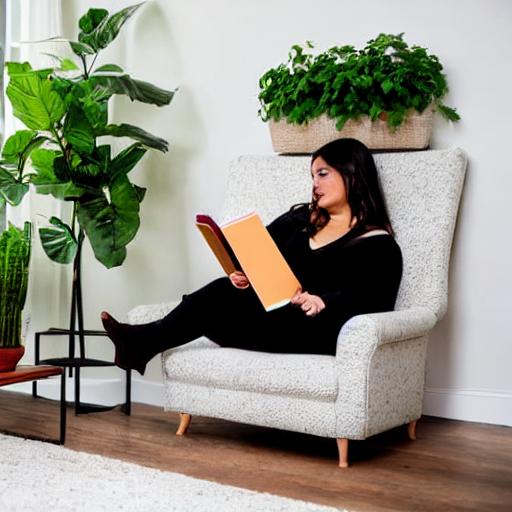} &
    \includegraphics[width=0.12\textwidth]{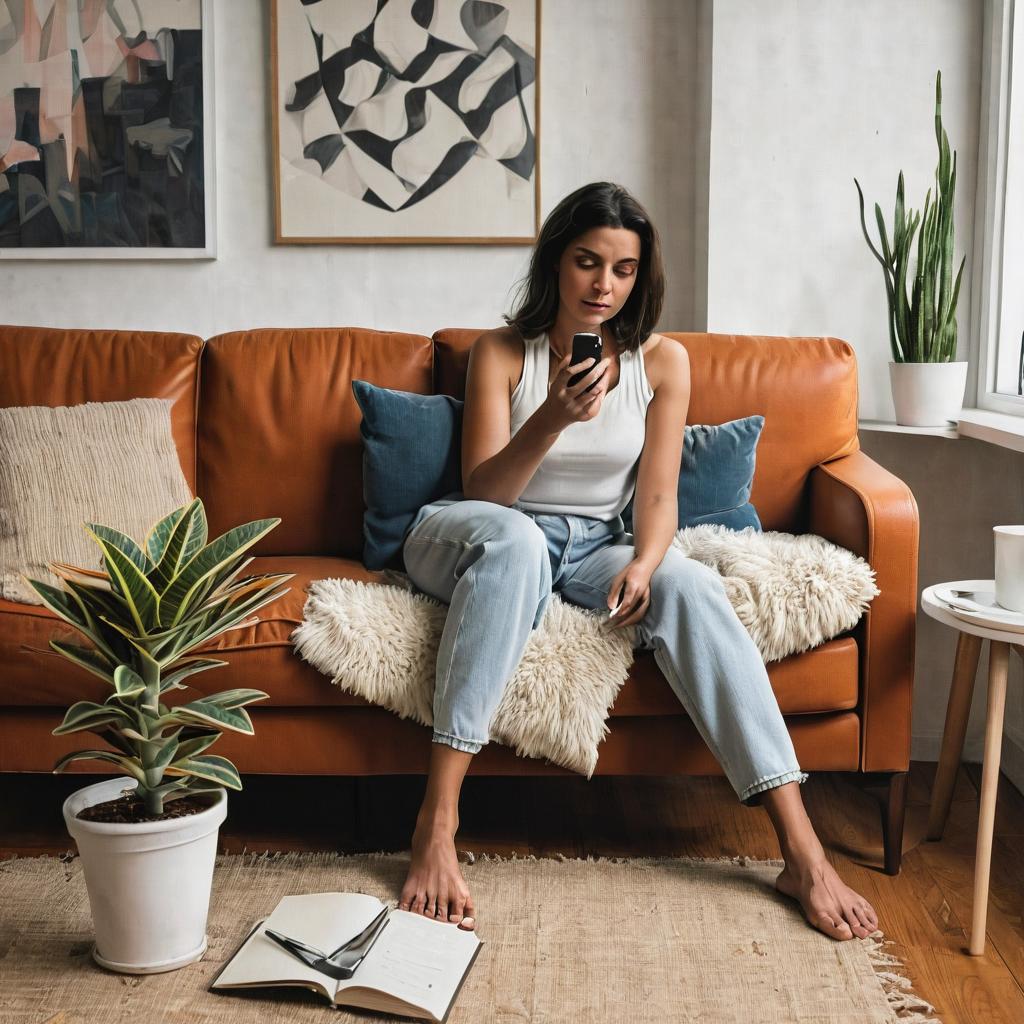} &
    \includegraphics[width=0.12\textwidth]{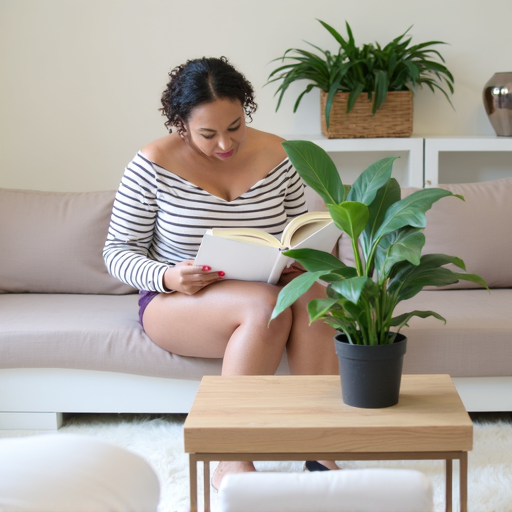} &
    \includegraphics[width=0.12\textwidth]{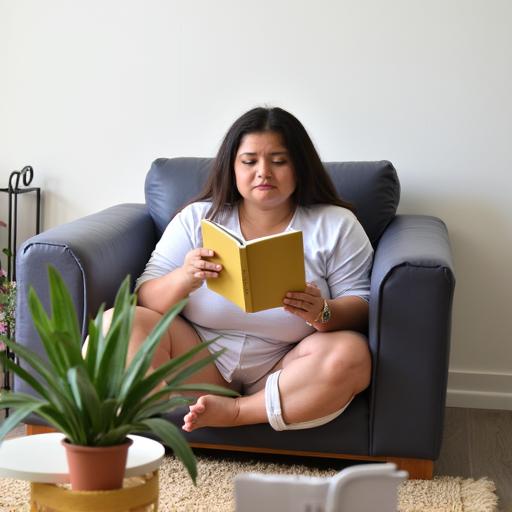} &
    \includegraphics[width=0.12\textwidth]{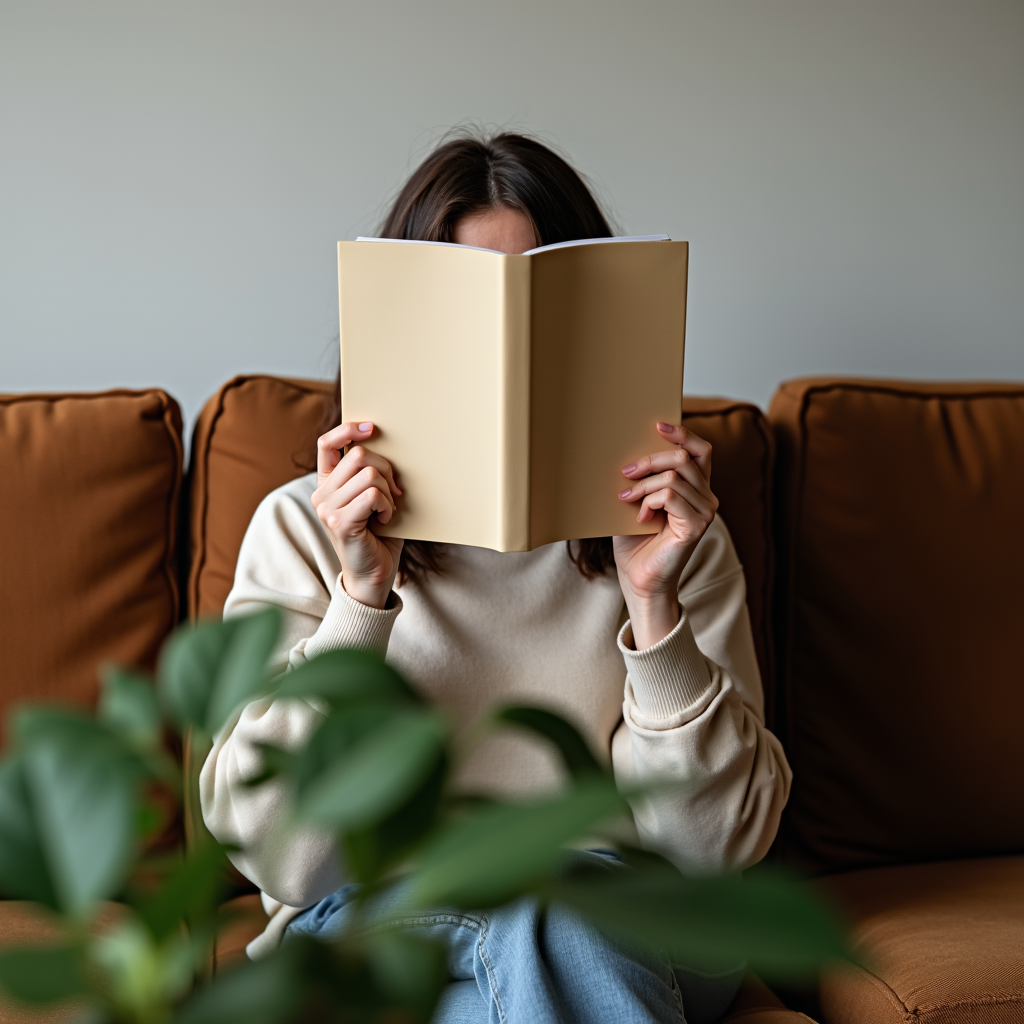} \\
    \multicolumn{8}{l}{\scriptsize \textit{Prompt: A \textcolor{blue}{woman} sitting on a \textcolor{green}{couch} reads a \textcolor{pink}{book} held up in her hands, while a \textcolor{yellow}{plant} sits in front of the couch.}} \\
    \addlinespace[3pt]

    \includegraphics[width=0.12\textwidth]{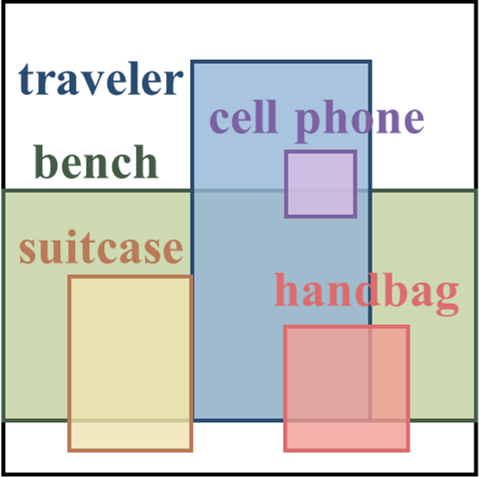} &
    \includegraphics[width=0.12\textwidth]{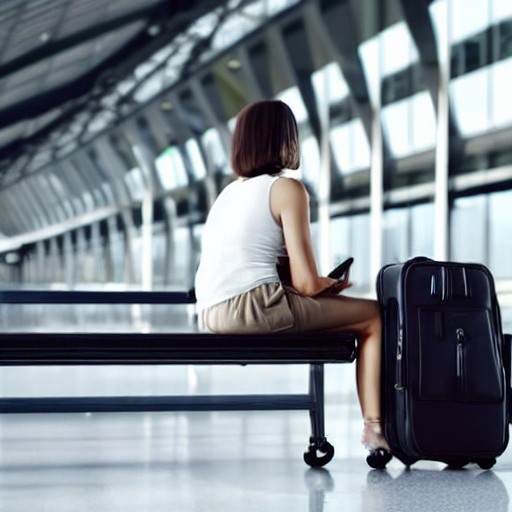} &
    \includegraphics[width=0.12\textwidth]{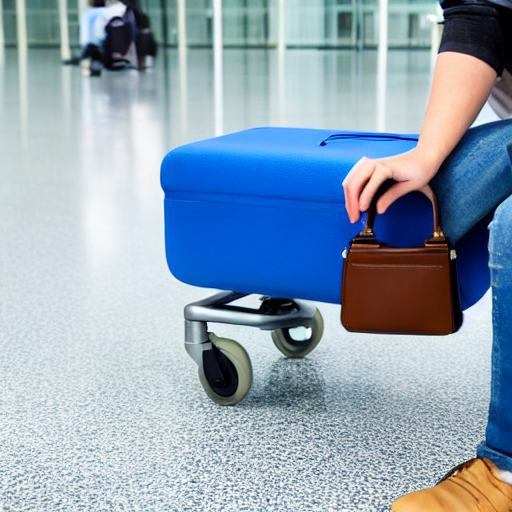} &
    \includegraphics[width=0.12\textwidth]{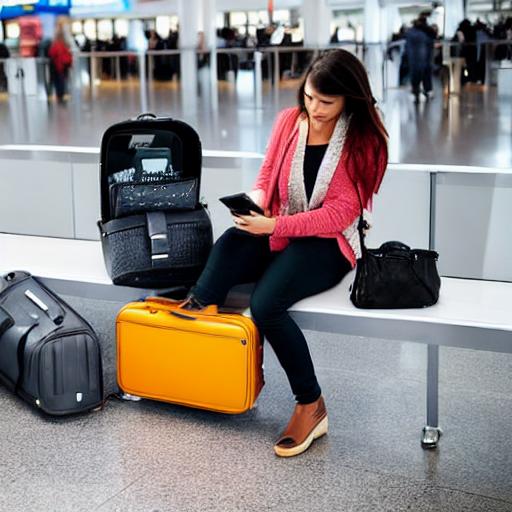} &
    \includegraphics[width=0.12\textwidth]{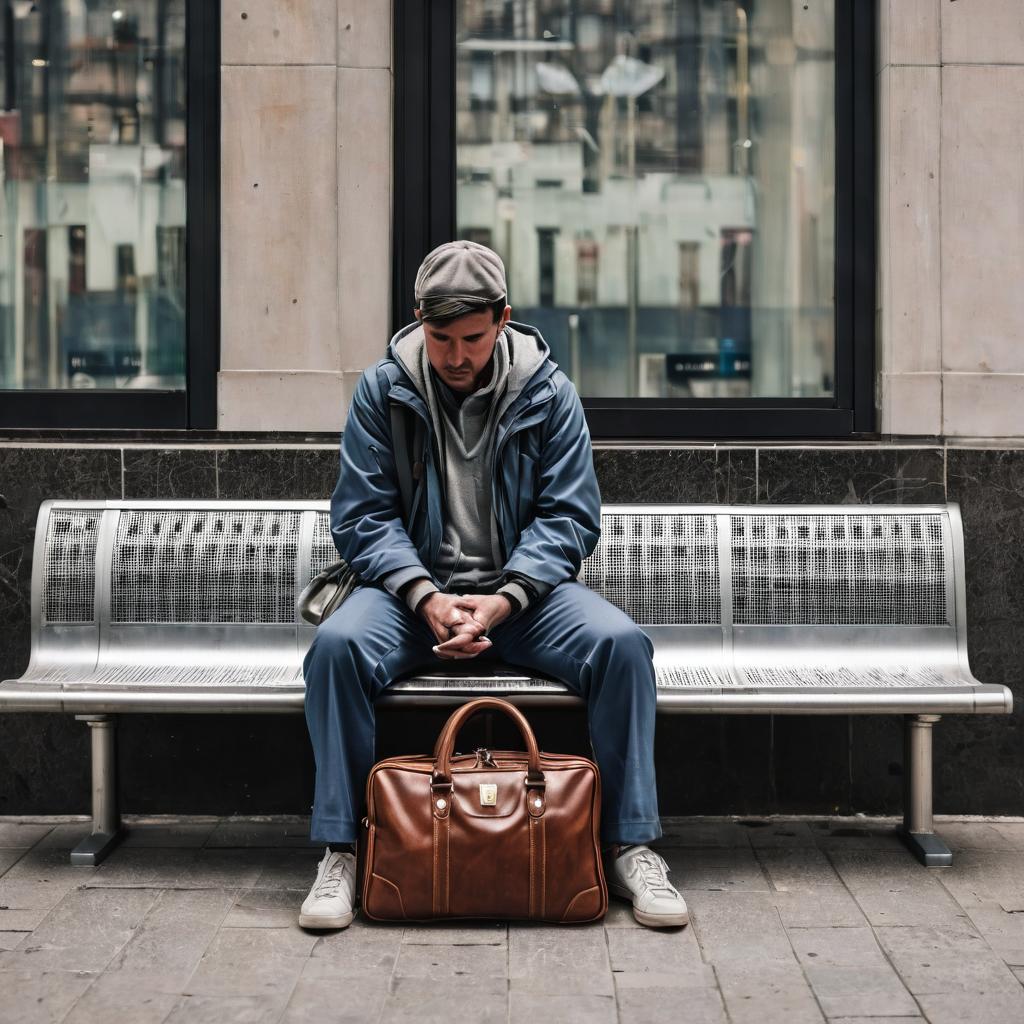} &
    \includegraphics[width=0.12\textwidth]{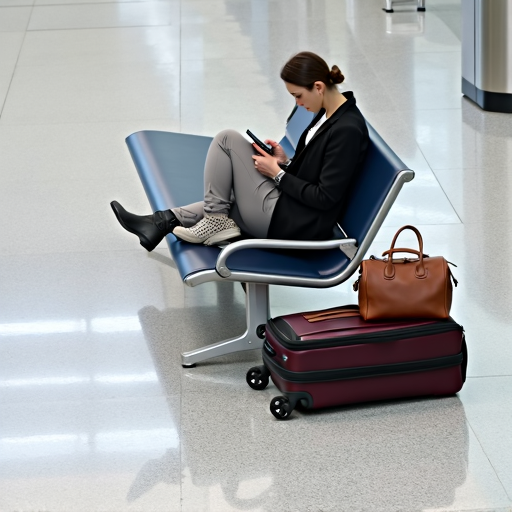} &
    \includegraphics[width=0.12\textwidth]{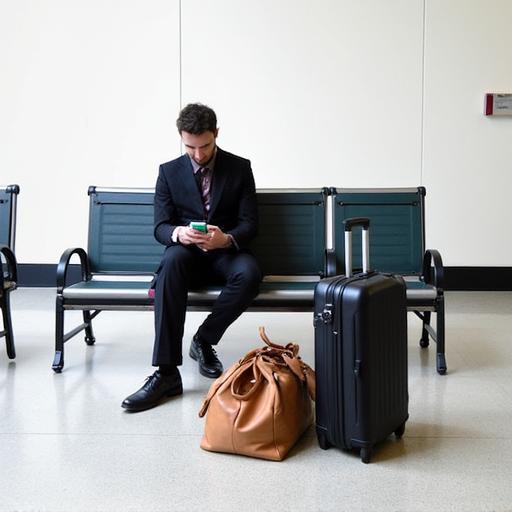} &
    \includegraphics[width=0.12\textwidth]{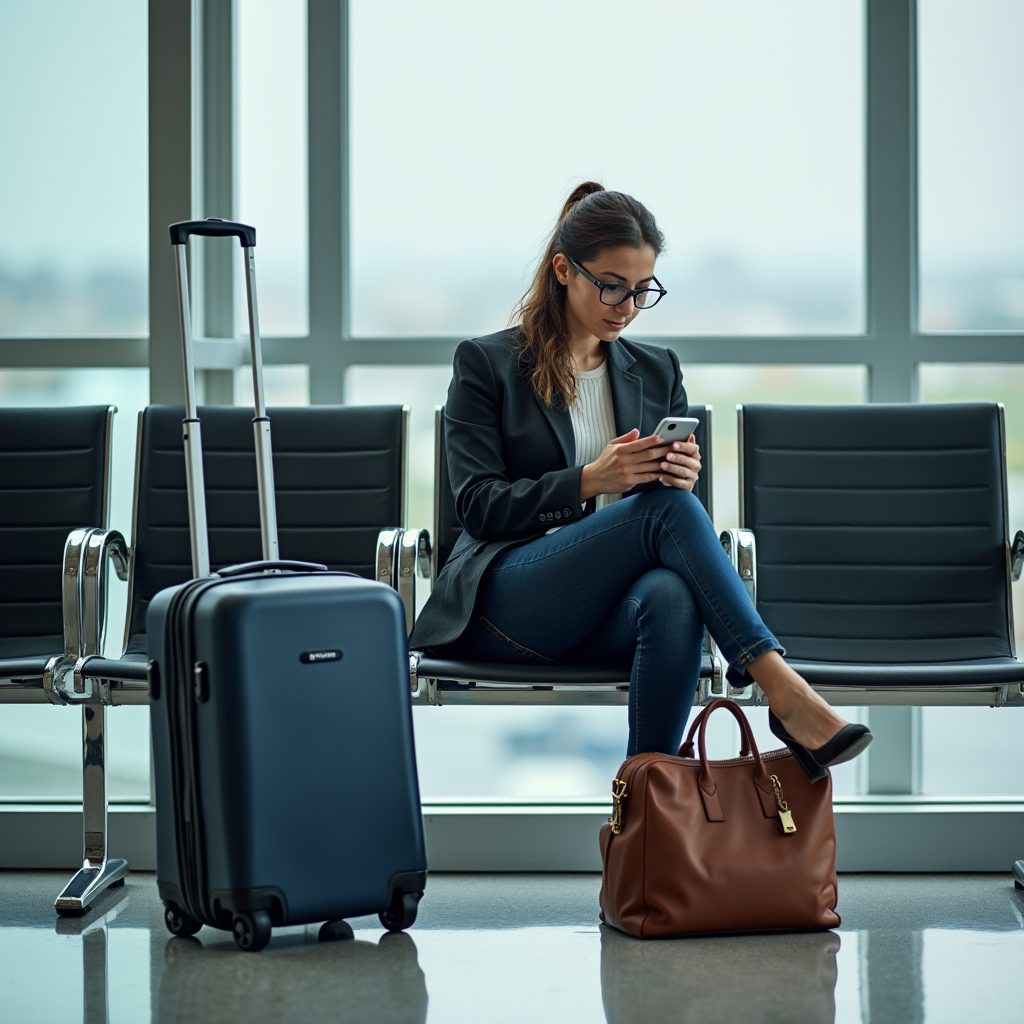} \\
    \multicolumn{8}{l}{\scriptsize \textit{Prompt: A \textcolor{blue}{traveler} sits on a metal airport \textcolor{green}{bench} looking at a \textcolor{purple}{cell phone}, with a large wheeled \textcolor{yellow}{suitcase} and a leather \textcolor{pink}{handbag} resting on the floor in front of them.}} \\

\end{tabular}

\caption{Qualitative comparisons on representative occlusion cases. Each row corresponds to one prompt with specified object ordering, and columns show the input layout, baselines, and DepthArb, respectively.
}
\label{fig:visualization}
\end{figure*}

\begin{table*}[!t]
\centering
\resizebox{\linewidth}{!}
{
\begin{tabular}{l|c|c|c|c|c|c|c|c}
\hline
\multirow{2}{*}{Method}
 & \multicolumn{3}{c|}{Layout Alignment (\%)} 
 & \begin{tabular}[c]{@{}c@{}}Text--Image\\Consistency\end{tabular}
 & \multicolumn{3}{c|}{Occlusion Quality} 
& \begin{tabular}[c]{@{}c@{}}Inference\\Time\end{tabular} \\
\cline{2-9}

 & mIoU-fg$\uparrow$ & mIoU-bg$\uparrow$ & mIoU-all$\uparrow$
 & CLIP Score$\uparrow$
 & FOCR (\%)$\uparrow$ & BOR$\uparrow$ & FBS$\uparrow$
 & Time (s)$\downarrow$\\
\hline
SDXL & 30.71 & 42.03 & 36.73 & 33.44 & 56.20 & 24.42 & 69.7 & 3.81 \\
FLUX & 36.98 & 50.54 & 42.48 & \textbf{34.01} & 61.22 & 24.91 & 71.5 & 6.57\\
Layout Guidance & 27.32 & 43.57 & 34.59 & 32.00 & 52.30 & 25.37 & 58.6 & 4.05 (+130\%)\\
R\&B & 55.15 & 59.12 & 57.88 & 31.60 & 70.27 & 25.86 & 73.3  & 26.89 (+684\%)\\
VODiff & 56.01 & 58.12 & 57.98 & 32.39 & 73.55 & 25.17 & 76.1& 8.85 (+48\%)  \\
LaRender & 56.18 & 60.26 & 58.33 & 32.17 & 66.93 & 25.82 & 74.8 & 6.03 (+171\%)  \\
\hline
Ours w/o LC & 32.00 & 47.09 & 39.13 & 33.16 & 69.18 & 24.51 & 71.2 & 4.77 (+25\%)\\
Ours w/o OCE & 55.40 & 59.10 & 58.07 & 33.07 & 80.20 & 25.69 & 87.5 & 4.79 (+26\%)\\
Ours w/o SCC & 53.65 & 57.43 & 56.12 & 32.30 & 75.40 & 24.73 & 77.0 & 4.74 (+24\%)\\
Ours w/o AAM & 53.12 & 56.98 & 55.76 & 32.23 & 78.56 & 24.66 & 74.4 & 4.81 (+26\%)\\

\hline
Ours (full) + SDXL & \underline{56.28} & \underline{61.17} & \underline{59.96} & 33.41 & \underline{81.77} & \textbf{25.98} & \underline{88.1} & \underline{4.85 (+27\%)} \\
Ours (full) + FLUX & \textbf{57.53} & \textbf{61.28} & \textbf{60.68} &  \underline{33.58} & \textbf{83.79} & \underline{25.91} & \textbf{88.7} & \textbf{7.48 (+14\%)}\\
\hline
\end{tabular}
 }
 \caption{Quantitative comparison of base models, training-free layout-guided baselines, four SDXL-based ablation variants, and the full DepthArb models on OcclBench. Percentage overheads are computed relative to each method's corresponding base model. \textbf{Bold} and \underline{underline} denote the best and the second best results, respectively.}
 \label{tab:table}
\end{table*}

\subsection{Benchmarks}
\paragraph{OverLayBench.}

To comprehensively assess robustness under complex spatial conditions, we evaluate on OverLayBench\cite{li2025overlaybench}. Unlike conventional text-to-image benchmarks that primarily focus on global fidelity, image aesthetics and semantic alignment \cite{gokhale2022benchmarking, huang2025t2i, saharia2022photorealistic, bakr2023hrs}, OverLayBench deliberately targets challenging scenarios characterized by large overlapping regions, dense spatial compositions, and minimal semantic distinction between adjacent objects. We leverage its difficulty stratification to demonstrate the superiority of our depth arbitration mechanism in maintaining object identities and boundary adherence across a broad spectrum of challenging occlusion scenarios.

\paragraph{OcclBench.}
Although OverLayBench covers dense bounding box intersections, it does not systematically evaluate multi-object interactions with strict depth hierarchies. We therefore introduce OcclBench, a benchmark for occlusion-aware generation with fine-grained depth stratification.
 
OcclBench comprises 310 manually verified prompts, each containing 2--6 objects with typically 2--4 (up to eight) occlusion pairs and continuous depths forming hierarchies of up to three levels. We derive common object categories from MS-COCO \cite{lin2014microsoft} and organize them into semantically coherent groups through physically plausible interactions (e.g., ``a bird perched on a branch''). To enable precise yet scalable construction, we employ a human-in-the-loop pipeline in which ChatGPT \cite{openai_gpt5} synthesizes descriptive prompts and spatial layouts, followed by manual verification for consistency with the textual intent. Detailed dataset construction and verification procedures are provided in the supplementary material.

\paragraph{Baselines.}
We compare DepthArb with the unguided SDXL and FLUX backbones and several training-free spatial-control baselines: BoxDiff \cite{xie2023boxdiff}, Layout Guidance \cite{chen2024training}, R\&B \cite{ICLR2024_6f61f25a}, VODiff \cite{liang2025vodiff}, and LaRender \cite{Zhan_2025_ICCV}. BoxDiff, Layout Guidance, and R\&B represent general box-constrained attention-guidance methods, whereas VODiff and LaRender provide stronger comparisons for scenes involving overlapping or ordered objects. For qualitative evaluation, we additionally include the training-based methods CreatiLayout \cite{zhang2025creatilayout} and InstanceAssemble \cite{xiang2026instanceassemble}, both of which are built on the FLUX backbone. All methods are evaluated using the same prompts and, where applicable, prescribed layouts.

\subsection{Evaluation Metrics}
We quantitatively evaluate model performance on OcclBench along three primary axes:
(1) \textbf{Layout Alignment}: We measure spatial adherence via the mean IoU (mIoU) between the prescribed bounding boxes and those detected by Grounding-DINO \cite{liu2024grounding}. Higher values indicate closer agreement with the input spatial constraints.
(2) \textbf{Text--Image Consistency}: The semantic correspondence between the generated image and its prompt is quantified using the standard CLIP score \cite{radford2021learning}.
(3) \textbf{Occlusion Quality}: To conduct a fine-grained analysis of occlusion relationships, we introduce a suite of three specialized metrics. First, the Foreground Occlusion Coverage Ratio (\textbf{FOCR}) assesses depth ordering by quantifying the area proportion within the detected bounding box intersection covered by the detected foreground bounding box from Grounding-DINO. Second, Background Object Recognizability (\textbf{BOR}) evaluates the semantic integrity of partially occluded entities using the CLIP similarity between their unoccluded regions and nominal class labels. Finally, Foreground--Background Separability (\textbf{FBS}) assesses boundary clarity and amorphous transitions (i.e., concept bleeding) using Gemini 3 Pro \cite{gemini_3_pro} for automated visual judgment. On 150 stratified images, FBS strongly correlates with mean human judgments (Spearman's $\rho=0.81$), with high inter-annotator reliability (Krippendorff's $\alpha=0.84$).

\begin{table*}[!t]
\centering
\resizebox{\linewidth}{!} 
{
\setlength{\tabcolsep}{4pt} 
\renewcommand{\arraystretch}{1.1}
\begin{tabular}{l | c c c c c c | c c c c c c} 
\hline
\multirow{2}{*}{Method}
& \multicolumn{6}{c|}{\textbf{OverLayBench-Regular}} 
& \multicolumn{6}{c}{\textbf{OverLayBench-Complex}} \\

\cline{2-13}
& mIoU$\uparrow$ & O-mIoU$\uparrow$ & SR$_E$$\uparrow$ & SR$_R$$\uparrow$ & CLIP$_\text{G}$$\uparrow$ & CLIP$_\text{L}$$\uparrow$ 
& mIoU$\uparrow$ & O-mIoU$\uparrow$ & SR$_E$$\uparrow$ & SR$_R$$\uparrow$ & CLIP$_\text{G}$$\uparrow$ & CLIP$_\text{L}$$\uparrow$ \\
\hline
BoxDiff & 19.40 & 5.33 & 37.58 & 71.81 & \underline{36.50} & 19.97 & 20.02 & 5.21 & 33.52 & 76.41 & \textbf{36.92} & 19.91 \\
Layout Guidance & 15.81 & 4.51 & 44.94 & 72.76 & 31.76 & 20.05 & 16.34 & 4.01 & 37.76 & 75.53 & 32.75 & 19.72 \\
R\&B & 20.35 & 5.54 & 32.85 & 65.01 & 34.49 & 19.88 & 19.80 & 4.85 & 28.38 & 69.97 & 34.57 & 19.47 \\
VODiff & 42.80 & 19.67 & 50.85 & 69.03 & 30.27 & 22.33 & 42.48 & 20.88 & 43.67 & 70.20 & 32.98 & 20.58\\
LaRender & \underline{57.21} & 30.06 & \textbf{54.97} & \underline{75.40} & 28.67 & \underline{26.07} & 52.27 & \underline{25.07} & 45.08 & 74.62 & 26.85 & 24.43 \\
\hline
Ours + SDXL& 53.74 & \underline{30.24} & 49.54 & 74.83 & 35.94 & 25.27 & \underline{53.02} & 24.53 & \underline{64.14} & \underline{77.02} & 36.31 & \underline{25.96} \\
Ours + FLUX& \textbf{58.63} & \textbf{31.89} & \underline{51.45} & \textbf{76.07} & \textbf{37.69} & \textbf{28.10} & \textbf{58.21} & \textbf{30.55} & \textbf{69.73} & \textbf{77.34} & \underline{36.73} & \textbf{26.13} \\
\hline
\end{tabular}
}
\caption{Comparison between training-free methods on OverLayBench \cite{li2025overlaybench}.
\textbf{Bold} and \underline{underline} denote the best and the second best results, respectively.}
\label{tab:overlaybench} 
\end{table*}

\section{Results}

\subsection{Qualitative Evaluation}


\begin{figure}
  \centering
  \includegraphics[width=\columnwidth]{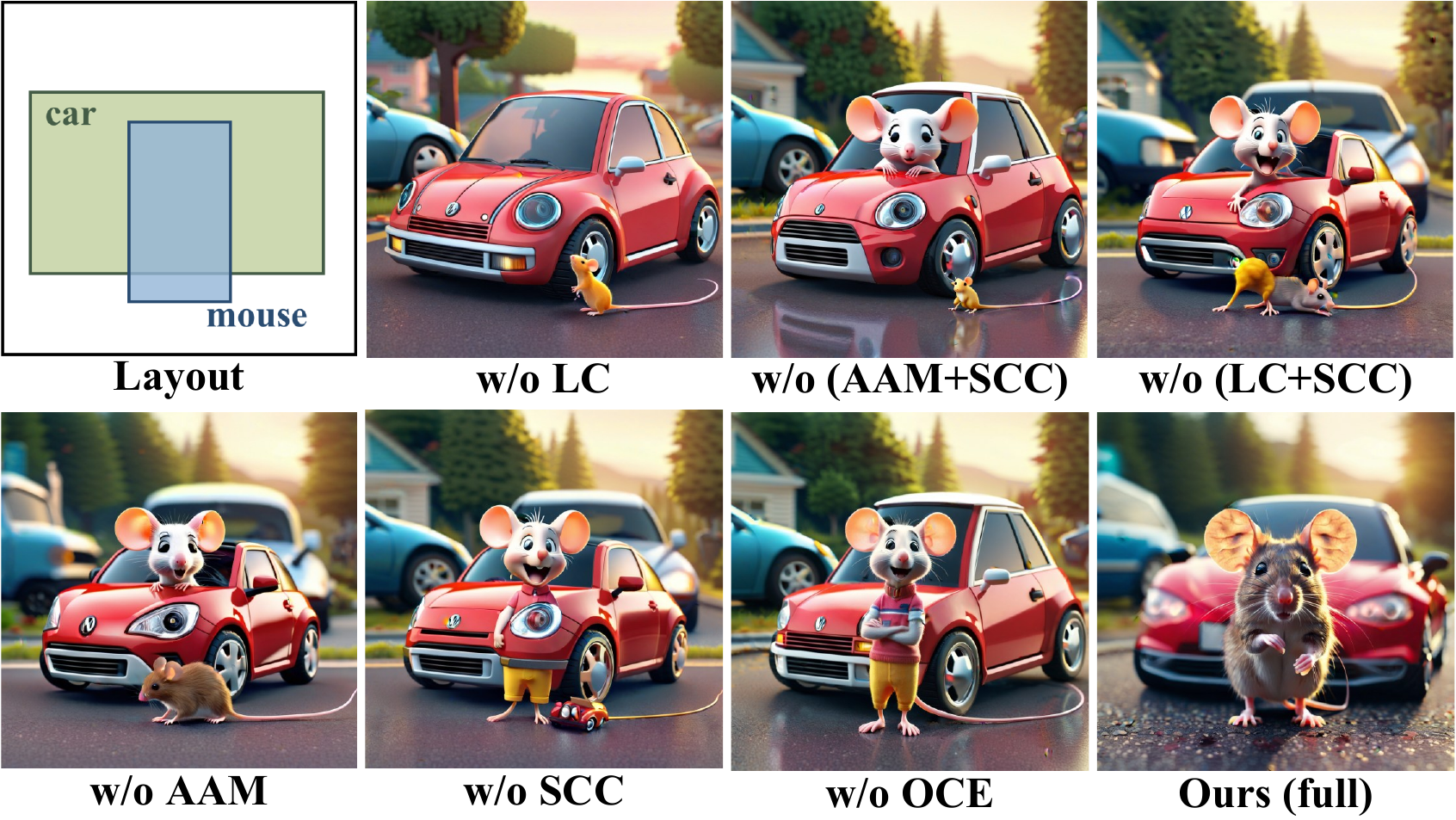}
  \caption{Visual ablation of component contributions. While the base model with only layout constraints exhibits concept mixing, the full DepthArb model produces clearer foreground--background separation in this example.
  }
  \label{fig:ablation}
\end{figure}

We qualitatively compare DepthArb against training-based \cite{zhang2025creatilayout, xiang2026instanceassemble} and training-free layout-guided baselines \cite{chen2024training, ICLR2024_6f61f25a, liang2025vodiff, Zhan_2025_ICCV} in Figure~\ref{fig:visualization}. Rather than relying on 2D placement alone or rendering ordered objects through separate stages, DepthArb preserves simultaneous generation and directly arbitrates feature competition, reducing concept missing and concept mixing in overlapping regions. 

\subsection{Quantitative Comparison}
As shown in Table~\ref{tab:table} and Table~\ref{tab:overlaybench}, DepthArb performs competitively on OcclBench and OverLayBench and obtains the best result on several metrics. Compared with the closest order-aware baseline VODiff in our evaluation, DepthArb improves FOCR from 73.55 to 81.77 and FBS from 76.1 to 88.1 on OcclBench; it also improves O-mIoU from 20.88 to 24.53 and SR$_E$ from 43.67 to 64.14 on OverLayBench-Complex. These results show that unified-trajectory guidance can be competitive with sequential visibility-layer synthesis. DepthArb integrated with FLUX gives the best results on most layout and occlusion metrics in these tables while introducing a 14\% inference overhead over the base model. Across both backbones, DepthArb consistently improves mIoU-all, FOCR, and FBS, which supports the transferability of the shared spatial-text interface without a substantial loss in global text-image alignment. DepthArb achieves the best CLIP$_\text{Global}$ score in the Regular setting and remains within 0.19 of the best result in the Complex setting. It also obtains the best CLIP$_\text{Local}$ scores in Table~\ref{tab:overlaybench}. These results suggest that the proposed spatial constraints maintain local semantic alignment in crowded scenes.

\subsection{Ablation Study}

We analyze DepthArb's components quantitatively (Table~\ref{tab:table}) and qualitatively (Figure~\ref{fig:ablation}). Removing LC substantially lowers spatial metrics (e.g., mIoU) and causes object drift, showing its importance for spatial alignment. Excluding AAM produces weaker depth ordering and more concept mixing between overlapping objects. Removing OCE reduces several layout and occlusion scores, including mIoU-all from 59.96 to 58.07 and FOCR from 81.77 to 80.20, consistent with a benefit from conflict-adaptive reweighting. Furthermore, removing SCC yields blurred occlusion boundaries and attention divergence, indicating that spatial compactness regularization stabilizes attention redistribution. Among the evaluated variants, the full model gives the strongest aggregate quantitative results and clearer qualitative separation.

\section{Conclusion}
In this paper, we introduce DepthArb, a training-free framework that reframes occlusion handling as conflict-adaptive attention arbitration. Unlike order-conditioned approaches that render discrete object layers through sequential denoising stages, DepthArb preserves simultaneous generation within a unified latent trajectory. Layout Confinement anchors object attention to target regions, while Occlusion Conflict Estimation modulates arbitration through a shared spatial conflict field. Attention Arbitration Modulation imposes depth-aware orthogonality to resolve inter-object competition, while Spatial Compactness Control stabilizes the redistributed attention. The same spatial--text interface applies to U-Net cross-attention and MMDiT joint attention. Evaluations on OcclBench and public benchmarks show improvements on several layout and occlusion metrics while maintaining competitive semantic alignment, without model retraining or object-wise denoising decomposition.

\bibliography{aaai2027}

\clearpage
\appendix
\setcounter{secnumdepth}{2}

\section*{Appendices}

\numberwithin{figure}{section}
\numberwithin{table}{section}
\numberwithin{equation}{section}

\input{appendix}

\end{document}

%% file: appendix.tex
\section{OcclBench}
\subsection{Dataset Statistics}

\begin{figure}[h]
    \centering
    \includegraphics[width=1\linewidth]{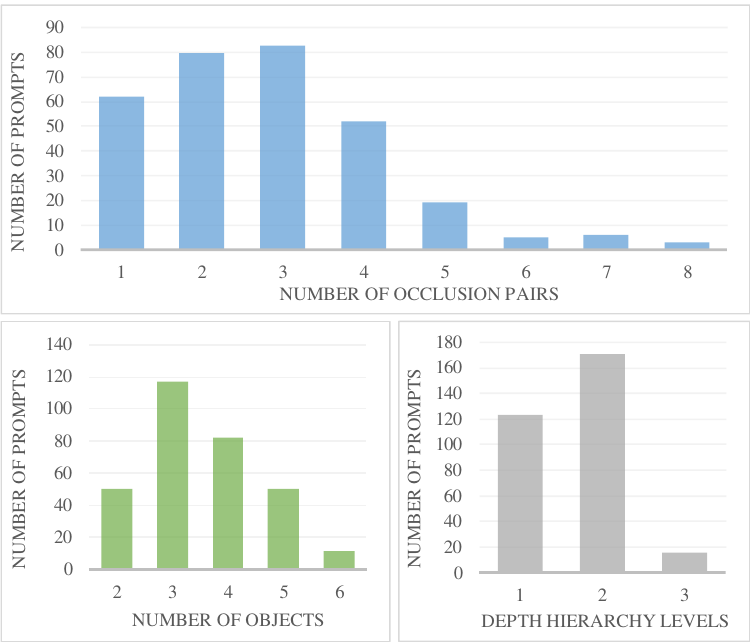}
    \caption{Distribution of the number of occlusion pairs (top), object counts (bottom left), and depth hierarchy levels (bottom right).}
    \label{fig:benchmark_statistics}
\end{figure}

Figure~\ref{fig:benchmark_statistics} summarizes the statistical characteristics of OcclBench. Each benchmark sample contains 2-6 objects with diverse spatial interactions, resulting in a broad distribution of pairwise occlusion relationships. Most samples involve 2-4 occlusion pairs, while more complex cases contain up to eight pairwise occlusions, providing challenging scenarios for multi-object occlusion reasoning. 

In addition, OcclBench explicitly annotates continuous depth hierarchies with up to three depth levels, enabling the evaluation of both simple foreground-background interactions and more complex layered occlusion structures. Overall, the benchmark covers a wide range of object compositions, occlusion complexities, and depth configurations, offering a comprehensive testbed for occlusion-aware text-to-image generation.

\subsection{Construction Pipeline}
\subsubsection{LLM Prompt}

We design a unified prompt template to automatically construct OcclBench from object categories sampled from MS-COCO~\cite{lin2014microsoft}. As shown in Figure~\ref{fig:occlbench_prompt}, ChatGPT~\cite{openai_gpt5} selects 2--6 object categories to synthesize a semantically coherent scene and simultaneously generates structured geometric annotations, including normalized bounding boxes, continuous depth values, and pairwise occlusion relationships. The generated annotations are subsequently refined through the human verification process described below.

\begin{figure*}
\centering
\begin{tcolorbox}[
    colback=gray!5,
    colframe=black,
    title=Prompt for OcclBench Construction,
    fonttitle=\bfseries,
    enhanced,
    sharp corners,
    width=\textwidth
]
\small

\textbf{Role:} You are an expert data annotator for constructing an occlusion-aware text-to-image benchmark. \\

\textbf{Task:} Given a candidate object pool sampled from the MS-COCO dataset, construct a realistic multi-object scene and generate a complete geometric annotation for the selected objects. \\

\textbf{Output JSON:}
\begin{itemize}
    \item A natural language prompt describing the generated scene.
    \item A list of the selected objects.
    \item A normalized bounding box for each selected object in the format \texttt{[y1, x1, y2, x2]}.
    \item A continuous depth value for each selected object.
    \item Pairwise occlusion relationships between overlapping objects.
\end{itemize}

\textbf{Requirements:}
\begin{itemize}
    \item Select \textbf{2--6} object categories from the input object pool.
    \item Construct a semantically coherent and physically plausible scene using all selected objects.
    \item Prefer object combinations that naturally produce meaningful occlusion relationships.
    \item Ensure realistic object interactions and physically consistent spatial arrangements.
    \item Generate normalized bounding boxes within the range $[0,1]$.
    \item Assign a \textbf{continuous depth} value in $[0,1]$ to each selected object rather than a discrete visibility order. Smaller values indicate objects closer to the camera, and the assigned depths should remain globally consistent with both the generated scene semantics and the spatial layout.
    \item Infer pairwise occlusion relationships directly from the generated layout and depth values using the format \texttt{[foreground, background]}.
    \item Ensure every selected object remains at least partially visible and avoid physically implausible layouts.
    \item Return only a valid JSON object.
\end{itemize}

\textbf{Example Output:}

\begin{verbatim}
{
  "id": "case_001",
  "prompt": "A little boy is playing the guitar.",
  "phrases": ["boy", "guitar"],
  "bboxes": [
      [0.10, 0.20, 1.00, 0.60],
      [0.40, 0.00, 0.60, 0.70]
  ],
  "depth": [0.70, 0.20],
  "occlusion_pairs": [
      ["guitar", "boy"]
  ]
}
\end{verbatim}

\end{tcolorbox}

\caption{The prompt template used to automatically construct OcclBench. Given a candidate object pool sampled from MS-COCO, the LLM selects 2--6 object categories, synthesizes a semantically coherent scene, and generates geometric annotations including bounding boxes, continuous depth values, and pairwise occlusion relationships.}

\label{fig:occlbench_prompt}

\end{figure*}

\subsubsection{Human Verification}

To improve the physical consistency of OcclBench, all automatically generated layouts undergo manual verification. Annotators inspect each sample along three dimensions. First, they correct inconsistent depth relationships to ensure that the depth hierarchy agrees with both the textual prompt and the spatial layout. Second, they remove cases involving complete occlusion so that every object remains at least partially visible. Finally, they refine excessive overlaps, unrealistic object scales, and misaligned bounding boxes. Samples that cannot be corrected without altering the original textual intent are discarded. This process reduces ambiguous or physically implausible layouts and improves the reliability of the benchmark.


\subsection{Comparison with Existing Benchmarks}
Table~\ref{tab:benchmark_comparison} compares our proposed OcclBench with T2I-CompBench++~\cite{huang2025t2i}, VOBench~\cite{liang2025vodiff}, RealOcc~\cite{Zhan_2025_ICCV}, and BindBench~\cite{Chen_2026_CVPR}. OcclBench contains 310 manually verified prompts with up to six objects. Its continuous depth annotations and multi-object layouts complement these benchmarks by supporting fine-grained evaluation of complex occlusion relationships.

\begin{table*}
\centering
\begin{tabular}{lcccc}
\toprule
\textbf{Benchmark} &
\textbf{Scale} &
\textbf{Objects} &

\textbf{Depth Annotation} &
\textbf{Complex Multi-Object} 
 \\
\midrule
T2I-CompBench++ & 1000 & 2 & None & \xmark   \\
VOBench & 200 & 2--5 & Visibility Order  & $\triangle$  \\
RealOcc & 70 & 2--5 & None & $\triangle$    \\
BindBench & 200 & 3--5  & Visibility Order & \cmark  \\
\midrule
\textbf{OcclBench (Ours)} &  310 & \textbf{2--6} & \textbf{Continuous Depth} & \cmark \\
\bottomrule
\end{tabular}
\caption{Comparison of OcclBench with existing benchmarks for occlusion-aware text-to-image generation.}
\label{tab:benchmark_comparison}
\end{table*}

\subsection{FBS Metric and Human Validation}
\paragraph{Automated FBS Evaluation.}
Foreground-Background Separability (FBS) evaluates whether overlapping foreground and background objects remain structurally independent, with clear boundaries and limited concept or texture bleeding. We use Gemini 3 Pro~\cite{gemini_3_pro} to assign an image-level score from 0 to 100 under a fixed rubric: higher scores indicate sharper boundaries and less semantic leakage, whereas lower scores indicate increasingly severe fusion or amorphous transitions between objects. The complete evaluation prompt and scoring rubric are shown in Figure~\ref{fig:fbs_prompt}.

\begin{figure}[tb]
\centering
\begin{tcolorbox}[
    colback=gray!5, 
    colframe=black, 
    title=Prompt for Foreground-Background Separability (FBS) Metric,
    fonttitle=\bfseries,
    enhanced,
    sharp corners,
    width=\columnwidth
]
\small
\textbf{Role:} You are an expert computer vision evaluator reviewing generated images. \\
\textbf{Task:} Assess the Foreground Background Separability (FBS) for the provided set of images. FBS evaluates the boundary articulation and structural independence between overlapping objects. \\
\textbf{Evaluation Criteria:} \\
You must penalize any amorphous transitions, concept bleeding, or texture leakage between objects. Assess each image using the following rigorous scale:
\begin{itemize}
    \item 90 to 100: Perfect separation. Boundaries are exquisitely sharp. Zero semantic leakage or texture blending occurs between the foreground and background.
    \item 70 to 89: Good separation. Minor boundary ambiguity exists, but the physical interaction remains highly plausible and distinct.
    \item 50 to 69: Moderate entanglement. Noticeable concept bleeding occurs. Boundaries begin to dissolve, making depth ordering slightly ambiguous.
    \item 0 to 49: Severe failure. Objects are fused together with physically impossible transitions and extreme texture leakage.
\end{itemize}
\textbf{Instructions:} 
1. Analyze each provided image carefully against the evaluation criteria. 
2. Assign a specific integer score out of 100 for every single image. 
3. Calculate the precise average score across the entire set. 
4. Provide your final response in a structured JSON format containing the individual scores and the final average score.
\end{tcolorbox}
\caption{The detailed system prompt utilized for the automated Foreground-Background Separability (FBS) evaluation.}
\label{fig:fbs_prompt}
\end{figure}

\paragraph{Human Validation Protocol.}
We validate the automated FBS metric on 30 OcclBench prompts sampled to cover different levels of occlusion complexity. For each prompt, we evaluate one image from each of four baselines (Layout Guidance, R\&B, VODiff, and LaRender) and one from DepthArb with SDXL, yielding 150 images in total. Each image is independently rated by three blinded annotators. The evaluation interface presents only the generated image, while hiding the method identity and Gemini score. Image order is randomized, and outputs from different methods for the same prompt are not shown consecutively.

Annotators are instructed to evaluate only the designated overlapping regions, focusing on boundary clarity, structural independence, and concept or texture bleeding. They explicitly ignore overall aesthetics, global text-image alignment, and layout accuracy outside the overlapping regions. Ratings use a five-point ordinal scale: 1 indicates severe fusion or substantial leakage; 2 indicates major boundary ambiguity; 3 indicates distinguishable objects with noticeable artifacts; 4 indicates clear separation with only minor local artifacts; and 5 indicates clean separation with no visible mixing. Annotators may select N/A when one or both designated objects are absent and separability cannot be assessed. N/A ratings are treated as missing values rather than evidence of good separability in the primary analysis.

\paragraph{Aggregation and Agreement Analysis.}
For image $k$, we average its $n_k$ valid human ratings as
\begin{equation}
\bar{s}_k=\frac{1}{n_k}\sum_{r=1}^{n_k}s_{k,r},
\end{equation}
and linearly map the result to the same 0-100 range as the automated metric:
\begin{equation}
\mathrm{FBS}_{\mathrm{human},k}=25(\bar{s}_k-1).
\end{equation}
We compute ordinal Krippendorff's $\alpha$ from the original 1-5 ratings to assess inter-annotator reliability. We then measure image-level rank agreement between the Gemini and aggregated human scores using Spearman's correlation. The human ratings achieve high inter-annotator reliability ($\alpha=0.84$), and the automated FBS scores correlate strongly with the mean human judgments ($\rho=0.81$). These results support FBS as an automated approximation of human-perceived foreground-background separability, to be interpreted together with the complementary layout and recognizability metrics.

\section{Implementation Details}

Experiments are conducted on NVIDIA H100 80GB and NVIDIA A800 80GB GPUs, with each generation run executed on a single GPU. All image-generation experiments are conducted at a resolution of $1024 \times 1024$ with
a batch size of 1 and a fixed random seed of 445. SDXL~\cite{podell2023sdxlimprovinglatentdiffusion} is evaluated in
FP16, while FLUX~\cite{labs2025flux1kontextflowmatching} is evaluated in BF16. VAE decoding is performed in
FP32 for numerical stability.

For SDXL, we use an LMS scheduler with 50 denoising steps and a
scaled-linear noise schedule, where
$\beta_{\mathrm{start}}=0.00085$,
$\beta_{\mathrm{end}}=0.012$, and the number of training timesteps is
1,000. Explicit classifier-free guidance
(CFG)~\cite{ho2021classifierfree} is applied with a guidance scale of
7.5. Loss-based guidance is enabled during the first 30\% of the
denoising trajectory: AAM is applied during the first 20\%, while the
alignment and compactness constraints remain active for an additional
10\%. For FLUX, we use a Flow Matching Euler scheduler with 28
denoising steps, 1,000 training timesteps, and a scheduler shift of
3.0. The model's native guidance conditioning is applied with a scale
of 3.5. All loss-based guidance terms are enabled during the first
30\% of the FLUX denoising trajectory.

For each guided denoising step, we perform at most five latent
optimization iterations. The loss scale is set to 50, the stopping
threshold is set to 0.1, and the backward-guidance strength is set to
1.0. For Attention Arbitration Modulation, we set the base repulsion
weight to $\lambda_0=0.5$, the scaling factor to $\alpha=1.0$, and the
temperature to $\tau=1.0$. The loss weights are set to
$\lambda_{\mathrm{ortho}}=0.2$ and
$\lambda_{\mathrm{compact}}=0.5$.

\section{Hyperparameter Sensitivity Analysis}
\begin{figure}
  \centering
  \includegraphics[width=\columnwidth]{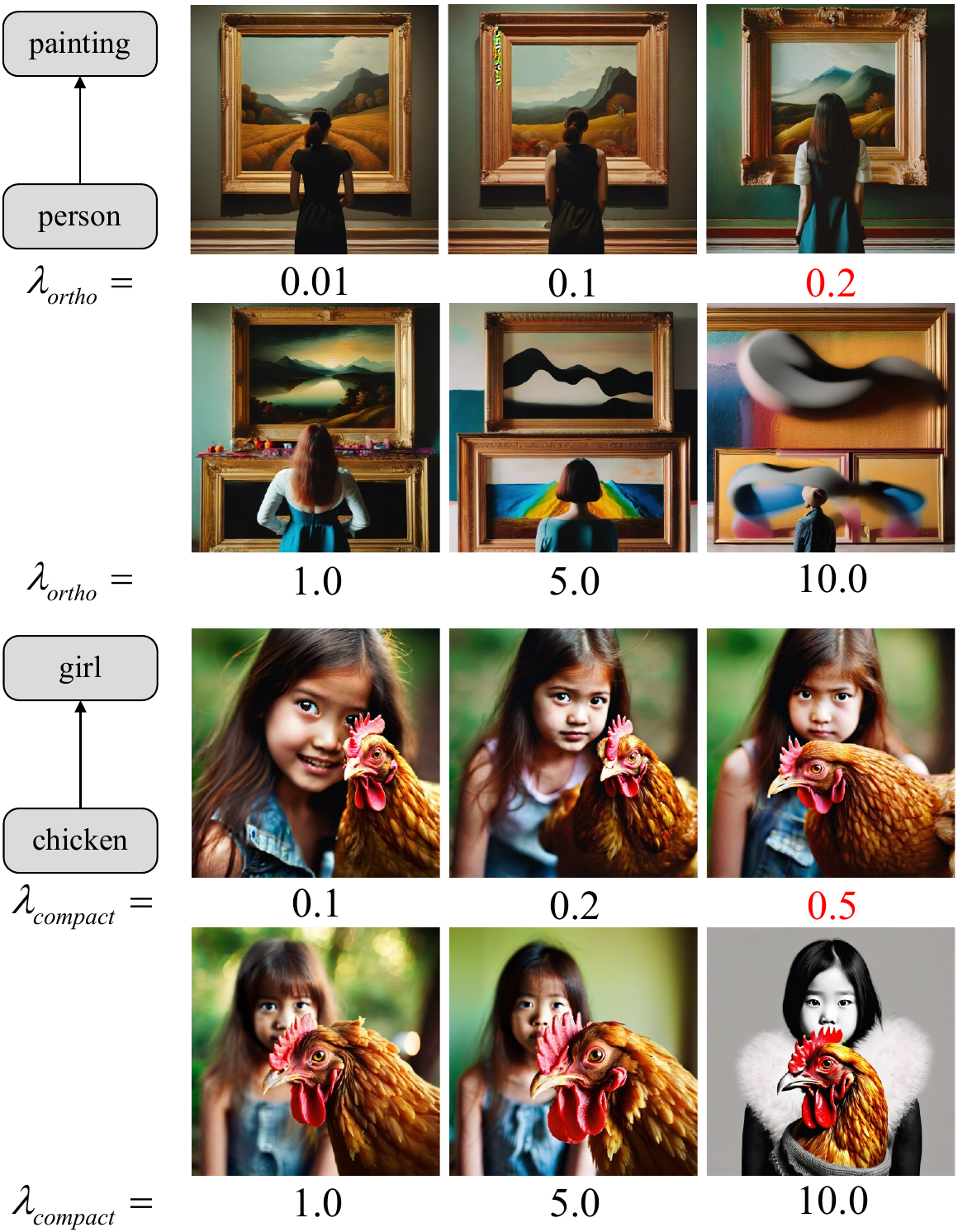}
  \caption{Analysis of key hyperparameters. We analyze $\lambda_{\mathrm{ortho}}$ (top) and $\lambda_{\mathrm{compact}}$ (bottom), with red values indicating default settings. Excessive $\lambda_{\mathrm{ortho}}$ disrupts background integrity, while extreme $\lambda_{\mathrm{compact}}$ values lead to either blurry boundaries or unnatural background style collapse.
  }
  \label{fig:sensitivity}
\end{figure}

We conduct the sensitivity analysis on the SDXL backbone, varying the orthogonality weight $\lambda_{\mathrm{ortho}}$ and compactness weight $\lambda_{\mathrm{compact}}$. The corresponding visual results are presented in Figure~\ref{fig:sensitivity}.

\paragraph{Effect of $\lambda_{\mathrm{ortho}}$.}
$\lambda_{\mathrm{ortho}}$ controls the suppression of background attention within foreground support (Figure~\ref{fig:sensitivity}, top). Excessively large values over-penalize spatial overlap, which can distort the background structure and introduce artifacts. We therefore set $\lambda_{\mathrm{ortho}}=0.2$ to balance foreground--background separation and background integrity.

\paragraph{Effect of $\lambda_{\mathrm{compact}}$.} 
$\lambda_{\mathrm{compact}}$ controls the spatial dispersion of object attention, with stronger compactness applied to nearer objects as defined in the main paper (Figure~\ref{fig:sensitivity}, bottom). Small values ($\le 0.2$) lead to fuzzy boundaries. In contrast, extreme values ($5.0 \sim 10.0$) can over-concentrate attention and disrupt the global context, causing background style collapse in some examples. We therefore use $\lambda_{\mathrm{compact}}=0.5$ to balance boundary clarity and natural integration.

\section{Quantitative Evaluation on T2I-CompBench++}

\begin{table*}
\centering

\begin{tabular}{@{} l cccccccc @{}}
\toprule
 & SDXL &  FLUX & Layout Guidance & R\&B & VODiff & LaRender & Ours+SDXL & Ours+FLUX\\
\midrule
UniNet $\uparrow$  & 0.354 & 0.383 & 0.327 & 0.330 & 0.361 & 0.378 & \underline{0.386} & \textbf{0.410}\\
CLIP Score $\uparrow$ & 31.45 & \underline{32.01} & 30.46 & 30.04 & 31.62 & 31.22 & 31.68 & \textbf{33.14} \\
\bottomrule
\end{tabular}
\caption{
Quantitative comparison on T2I-CompBench++.
  DepthArb with FLUX obtains the highest UniNet and CLIP scores among the evaluated methods. \textbf{Bold} and \underline{underline} denote the best and the second best results, respectively.
}
\label{tab:t2i_quantitative}
\end{table*}

To further evaluate the generalization and compositional generation capabilities of DepthArb, we use the 3D Spatial subset of T2I-CompBench++~\cite{huang2025t2i}. We compare DepthArb with the SDXL base model~\cite{podell2023sdxlimprovinglatentdiffusion}, the FLUX base model~\cite{labs2025flux1kontextflowmatching}, and four training-free methods: Layout Guidance~\cite{chen2024training}, R\&B~\cite{ICLR2024_6f61f25a}, VODiff~\cite{liang2025vodiff}, and LaRender~\cite{Zhan_2025_ICCV}.

As shown in Table~\ref{tab:t2i_quantitative}, we use UniNet to evaluate spatial layout accuracy and CLIP Score~\cite{radford2021learning} to measure overall text--image alignment. DepthArb with FLUX achieves the highest scores on both metrics among the evaluated methods. DepthArb with SDXL also achieves a higher UniNet score than LaRender and a slightly higher CLIP score than the base SDXL model. These results suggest that the improved spatial accuracy is not accompanied by a substantial degradation in global text--image alignment.

\section{More Qualitative Results}
\begin{figure}[tb]
  \centering
  \includegraphics[width=\columnwidth]{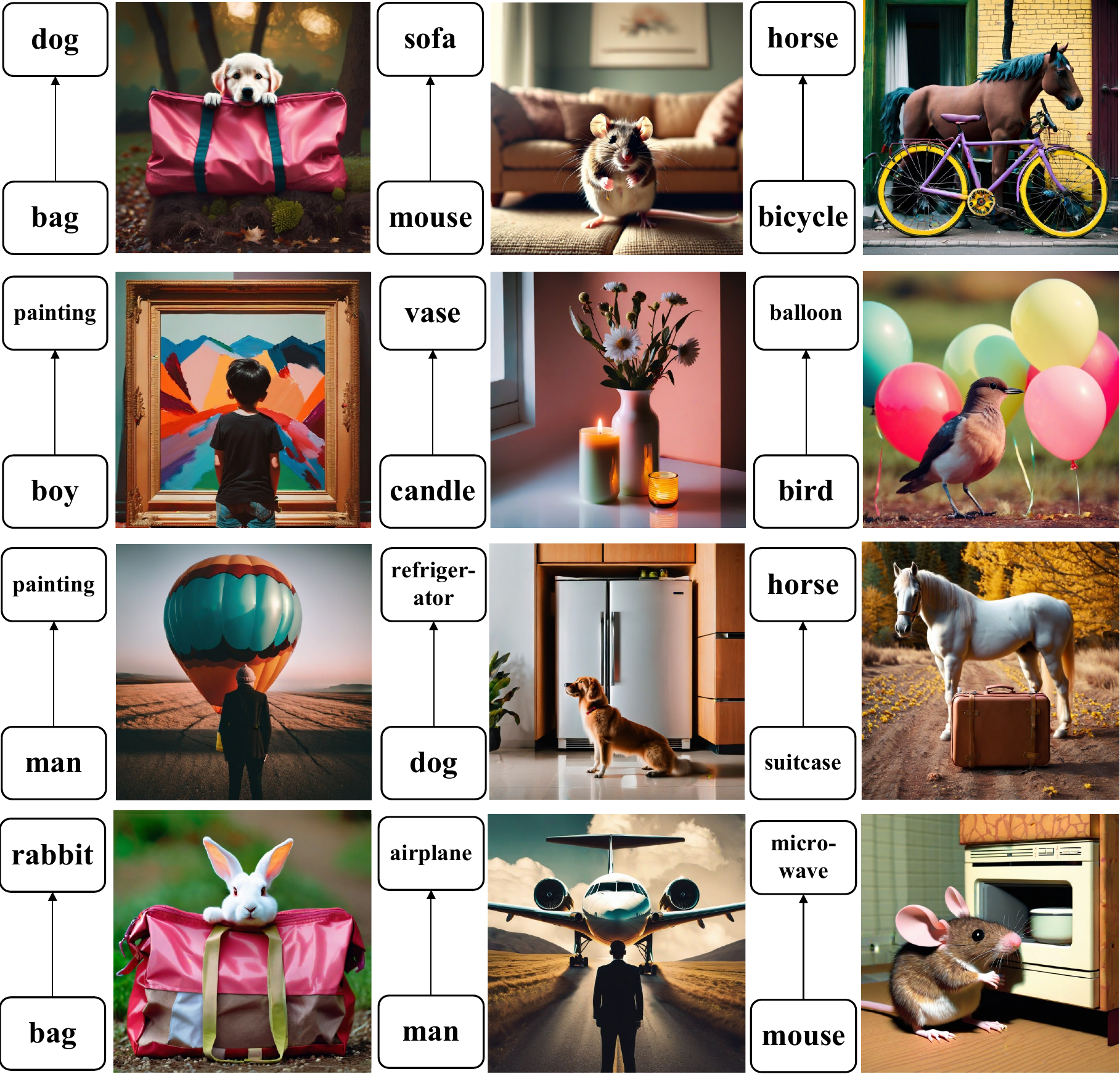}
  \caption{Qualitative results of DepthArb with SDXL on the T2I-CompBench++ benchmark.
  }
  \label{fig:t2i_qualitative_sdxl}
\end{figure}

\begin{figure}[tb]
  \centering
  \includegraphics[width=\columnwidth]{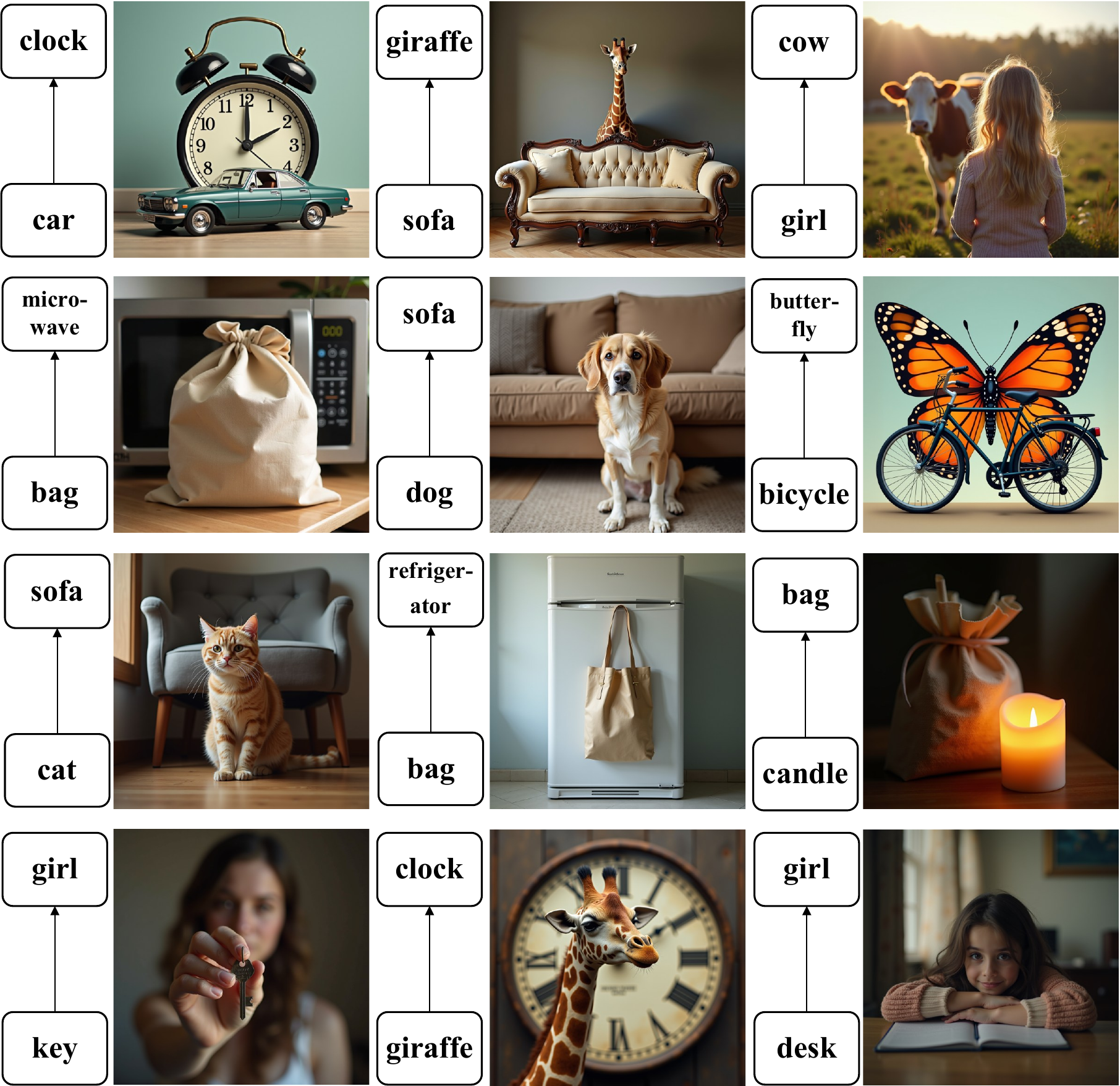}
  \caption{Qualitative results of DepthArb with FLUX on the T2I-CompBench++ benchmark.
  }
  \label{fig:t2i_qualitative_flux}
\end{figure}

We provide visual results using prompts from the T2I-CompBench++ dataset in Figure~\ref{fig:t2i_qualitative_sdxl} and Figure~\ref{fig:t2i_qualitative_flux}. In these examples, DepthArb follows the intended spatial and occlusion relationships while preserving plausible image appearance.

\begin{figure}
  \centering
  \includegraphics[width=\columnwidth]{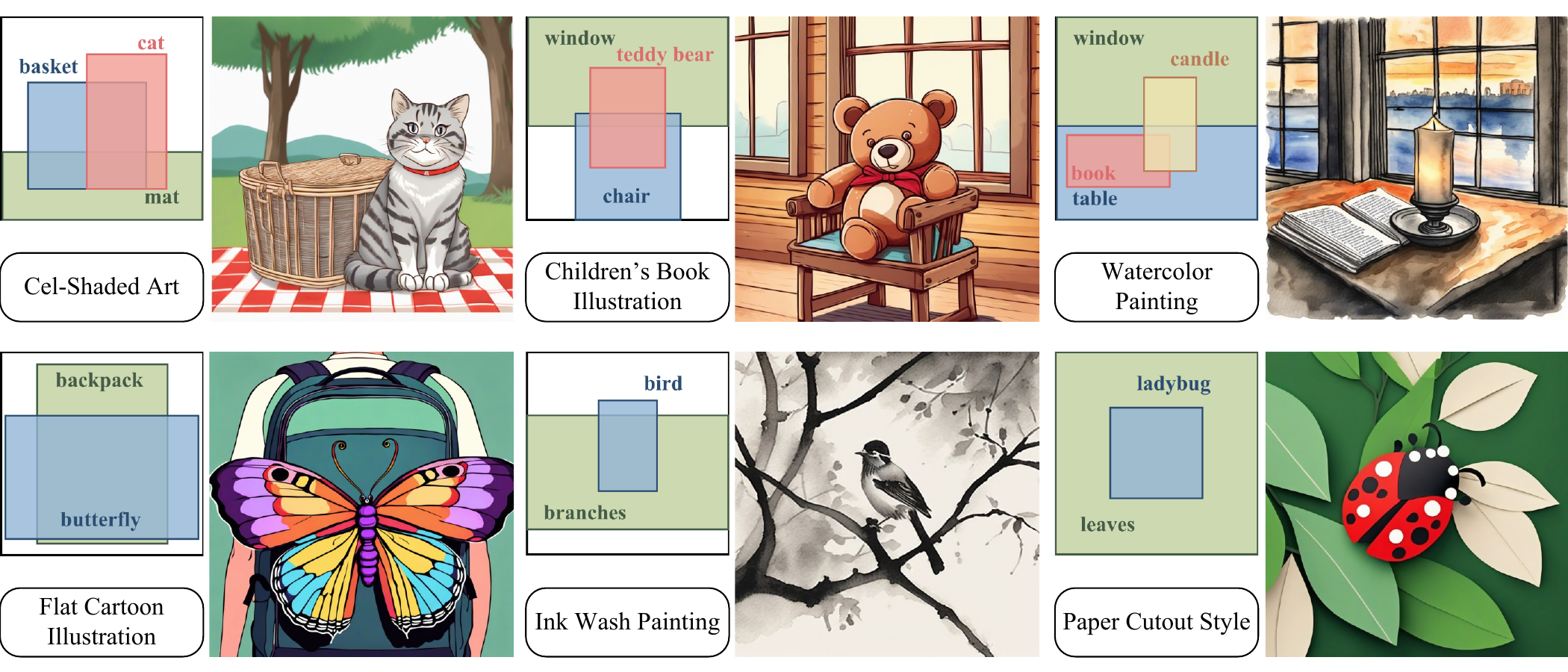}
  \caption{Qualitative results across diverse artistic styles. DepthArb follows the specified spatial layouts while preserving the intended styles in these examples.
  }
  \label{fig:style}
\end{figure}

To examine stylistic versatility, we apply DepthArb to several non-photorealistic domains. As shown in Figure~\ref{fig:style}, the method can follow complex layout and occlusion constraints across different artistic styles. These qualitative results suggest that the arbitration mechanism can operate without overriding the backbone's global stylization.

\section{User Study}
\label{sec:user_study}

\begin{table}
\centering

\begin{tabular}{lcc}
\toprule
 & Ours vs. SDXL & Ours vs. LaRender \\
\midrule
\shortstack[l]{Winning\\Rate (\%)} &
\smash{\raisebox{5pt}{71.3}} &
\smash{\raisebox{5pt}{73.1}} \\
\bottomrule
\end{tabular}
\caption{Pairwise user study results. Participants were asked to choose their preferred result between our method and each baseline independently.}
\label{tab:user_study}
\end{table}

We conducted a pairwise user study with 20 participants to evaluate human perceptual preference for DepthArb against SDXL and LaRender. For this study, DepthArb uses SDXL as its base model; therefore, all references to ``DepthArb'' or ``our method'' in this section denote the DepthArb+SDXL variant.

\paragraph{Evaluation Protocol.} 
Participants evaluated 15 prompt-image pairs from OcclBench for each baseline. In each trial, they selected their preferred result by jointly considering three aspects: (1) \textit{Overall visual quality}, (2) \textit{Text-image alignment}, and (3) \textit{Occlusion handling}. To prevent forced choices, a ``no noticeable difference'' option was provided. This setup yielded 300 responses per baseline comparison. We report the tie-excluded win rate, computed as $N_{\mathrm{ours}}/(N_{\mathrm{ours}}+N_{\mathrm{baseline}})$, with ``no noticeable difference'' responses excluded from the denominator.

\begin{figure}[tb]
  \centering
  \includegraphics[width=\columnwidth]{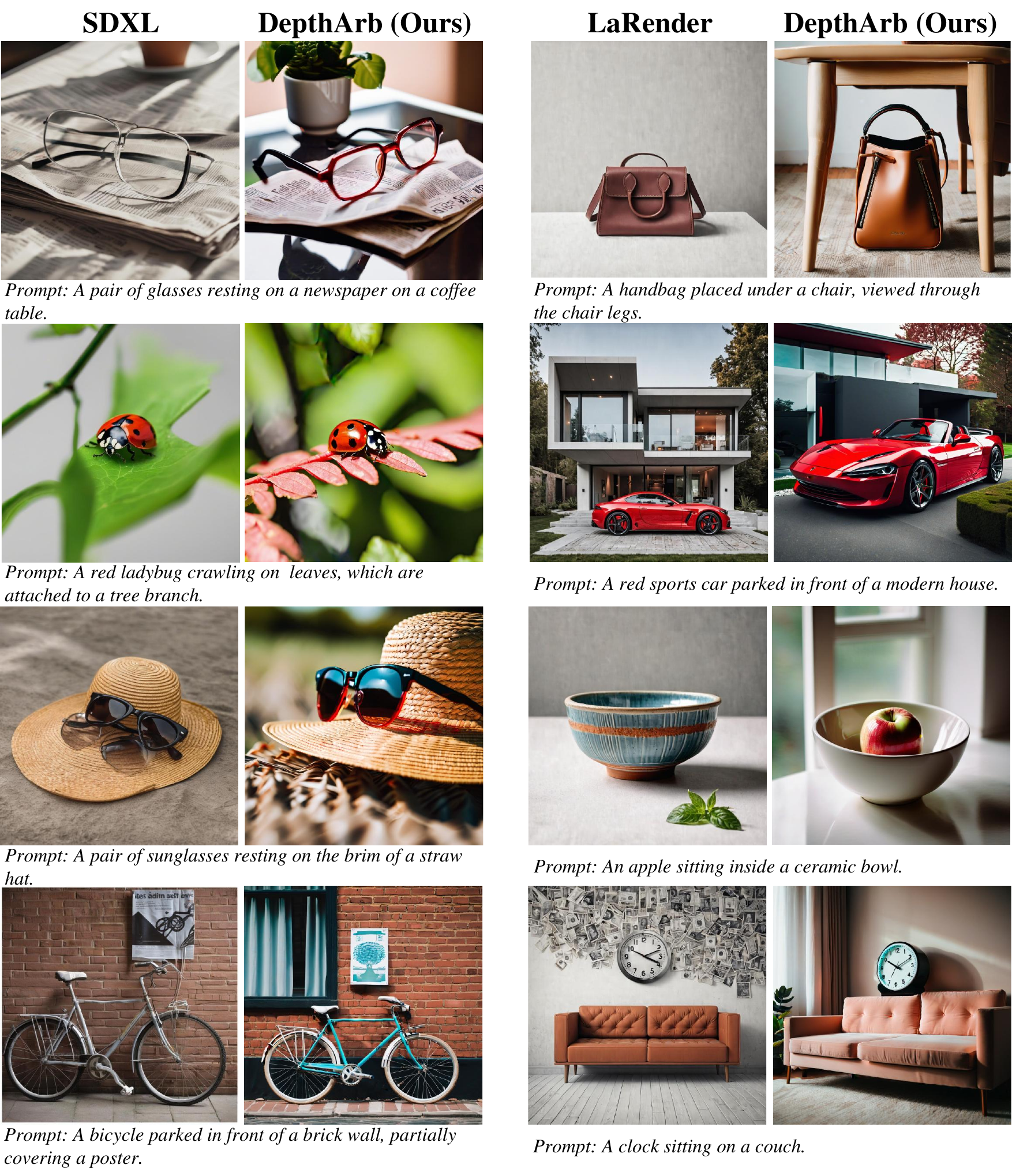}
  \caption{Example image pairs used in our user study. Participants compared DepthArb against SDXL (left) and LaRender (right). In these examples, DepthArb follows the specified spatial relationships more closely than the corresponding baseline.
  }
  \label{fig:userstudy_image}
\end{figure}

\paragraph{Results.} 
As detailed in Table~\ref{tab:user_study}, DepthArb achieves tie-excluded win rates of $71.3\%$ against SDXL and $73.1\%$ against LaRender. Representative examples in Figure~\ref{fig:userstudy_image} illustrate cases in which the baselines misinterpret spatial prepositions or fail to preserve the requested object relations, whereas DepthArb follows these constraints more closely. The preferences reflect the joint criterion of visual quality, text--image alignment, and occlusion handling used in the study.

\section{Limitations and Societal Impacts}

\begin{figure}[tb]
  \centering
  \includegraphics[width=\columnwidth]{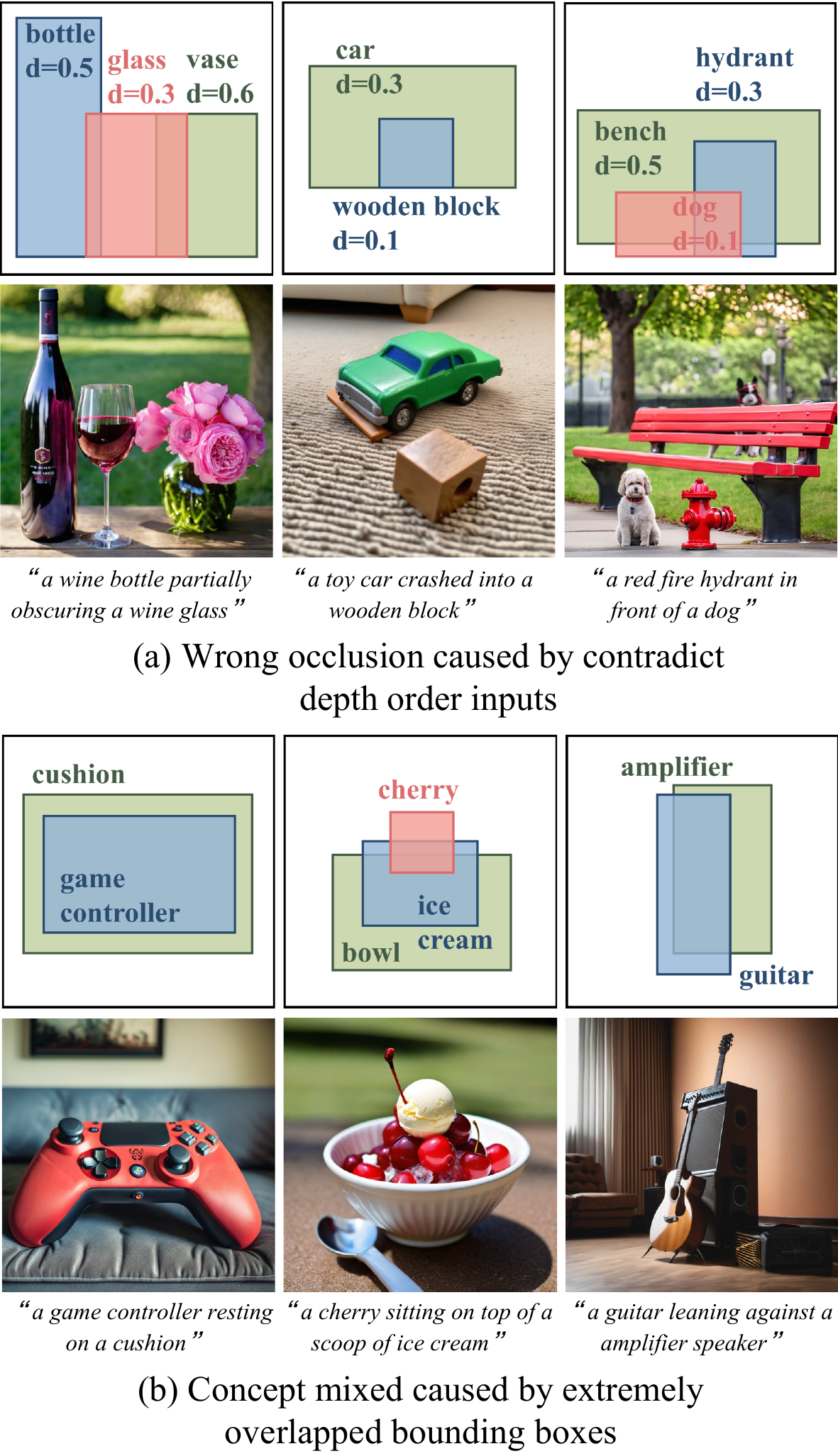}
  \caption{Representative failure cases of DepthArb.}
  \label{fig:limits}
\end{figure}

While DepthArb demonstrates strong performance, we identify two primary limitations, as illustrated in Figure~\ref{fig:limits}. First, our framework may falter when a prescribed depth order directly contradicts the model's semantic priors; in such cases, the model's inherent biases can override the arbitration mechanism, leading to an inversion of the intended hierarchy. Second, concept bleeding can still occur between visually similar objects or in regions of extreme overlap, where entangled token activations result in ambiguous boundaries. Addressing these edge cases will likely require more advanced semantic disentanglement techniques, which we leave as a direction for future research.

By enabling training-free spatial control, DepthArb provides users with finer control over complex spatial compositions, which may benefit creative applications, education, and design by reducing trial-and-error during generation. However, this ability also increases the risk of synthesizing deceptive or malicious content, such as deepfakes involving sensitive interactions between entities. We emphasize the necessity of deploying our framework responsibly and encourage the integration of origin-tracing watermarks to prevent potential misuse.